\documentclass[10pt,twocolumn,letterpaper]{article}

\usepackage{iccv}              

\definecolor{iccvblue}{rgb}{0.21,0.49,0.74}
\usepackage[pagebackref,breaklinks,colorlinks,allcolors=iccvblue]{hyperref}

\usepackage{wrapfig}
\usepackage{cuted}
\usepackage{multirow} 
\usepackage{graphicx}
\usepackage{tcolorbox}
\tcbuselibrary{breakable}

\def\paperID{13503} 
\def\confName{ICCV}
\def\confYear{2025}

\title{Multi-Object Sketch Animation by Scene Decomposition and Motion Planning}

\author{
    Jingyu Liu, 
    Zijie Xin,
    Yuhan Fu,
    Ruixiang Zhao,
    Bangxiang Lan,
    Xirong Li\footnotemark[2]\\
    \\
    Renmin University of China \\
    \url{https://rucmm.github.io/MoSketch}
}

\begin{document}
\maketitle
\begin{strip}
\begin{minipage}{\textwidth}\centering
\vspace{-35pt}
\includegraphics[width=\textwidth]{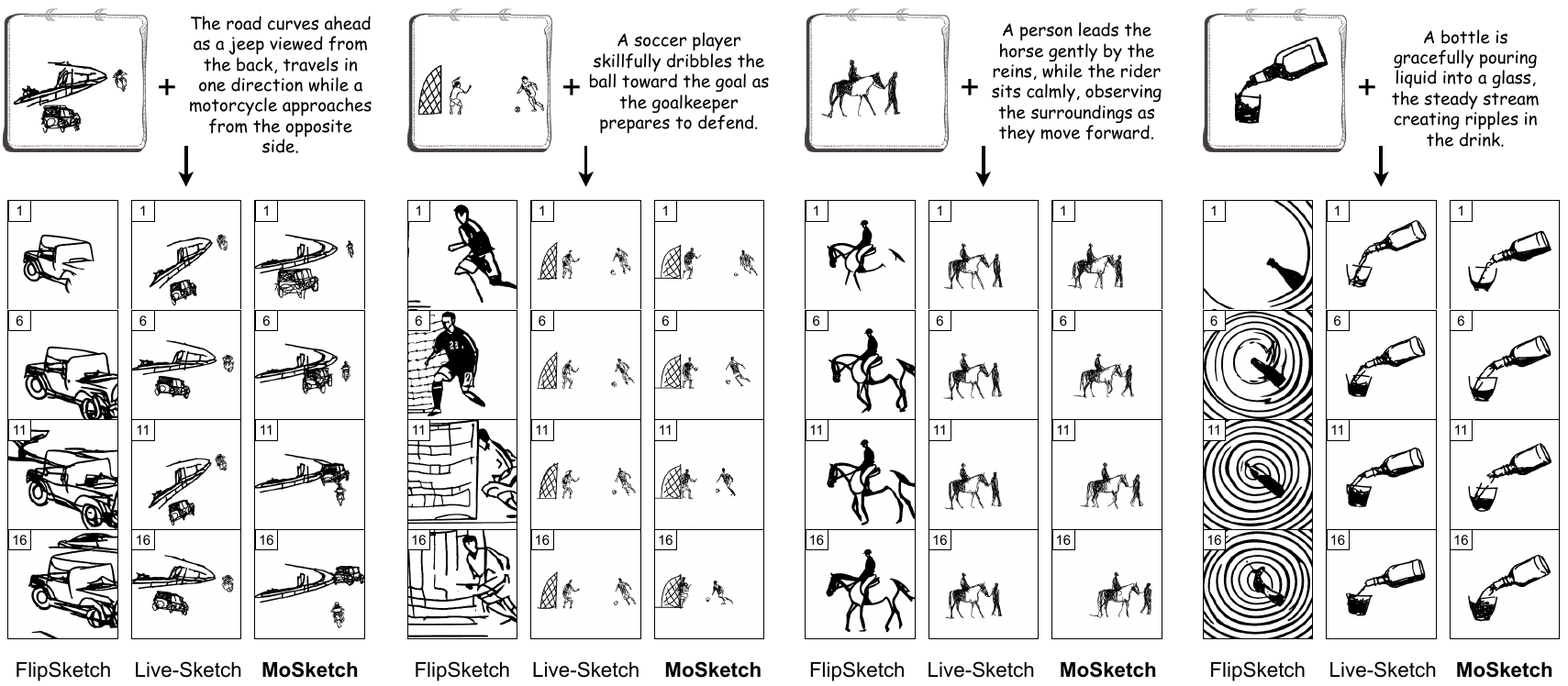}
\captionof{figure}{\textbf{Multi-object sketch animation}. State-of-art, \ie  FlipSketch \cite{bandyopadhyay2025flipsketch} and Live-Sketch \cite{gal2024breathing}, targeting at \emph{single}-object sketch animation,   struggles to animate \emph{multi}-object sketches. FlipSketch fails to preserve visual appearance, while Live-Sketch struggles with modeling complex motion, resulting in random shaky, nearly motionless animation, or even plausibility violation (\eg \emph{the liquid in the bottle increases with pouring}). We propose \textbf{MoSketch}, an iterative optimization based and thus training-data free method, to overcome these issues. 
}
\label{fig:intro}
\end{minipage}
\end{strip}

\renewcommand{\thefootnote}{\fnsymbol{footnote}} 
\footnotetext[2]{Corresponding author (xirong@ruc.edu.cn)}

\begin{abstract}

Sketch animation, which brings static sketches to life by generating dynamic video sequences, has found widespread applications in GIF design, cartoon production, and daily entertainment. While current methods for sketch animation perform well in single-object sketch animation, they struggle in \emph{multi}-object scenarios. By analyzing their failures, we identify two major challenges of transitioning from single-object to multi-object sketch animation: object-aware motion modeling and complex motion optimization. For multi-object sketch animation, we propose MoSketch based on iterative optimization through Score Distillation Sampling (SDS) and thus animating a multi-object sketch in a training-data free manner. To tackle the two challenges in a divide-and-conquer strategy, MoSketch has four novel modules, i.e., LLM-based scene decomposition, LLM-based motion planning, multi-grained motion refinement, and compositional SDS. Extensive qualitative and quantitative experiments demonstrate the superiority of our method over existing sketch animation approaches. MoSketch takes a pioneering step towards multi-object sketch animation, opening new avenues for future research and applications.

\end{abstract}
\section{Introduction}\label{sec:intro}
Sketch animation brings sketches composed of strokes to life by generating dynamic video sequences, making it widely used in GIF design, cartoon production, and daily entertainment~\cite{ha2018neural, su2018live, wei2024real}. We aim to animate a \emph{multi}-object sketch \wrt a specific textual instruction, see \cref{fig:intro}.

\begin{table} [htbp!]
\centering
\resizebox{\linewidth}{!}{
\begin{tabular}{lcccc}
\toprule
\multirow{2}*{\textbf{Method}} & \textbf{Sketch}  & \textbf{Object-aware} & \textbf{Training} & \textbf{Optimization} \\
                              &  \textbf{Representation} & \textbf{Motion Modeling} & \textbf{Data} & \textbf{Method}  \\
\midrule
Live-Sketch   & Vector & No & No & SDS \\
FlipSketch       & Raster & No & Yes & -- \\
\emph{MoSketch}      & Vector & Yes & No & \emph{Compositional} SDS \\
\bottomrule
\end{tabular}}
\caption{\textbf{Key properties of current methods}.}
\label{tab:intro}
\end{table}

Early methods for sketch animation require manual intervention \cite{davis2008k, yin2010sketch, yu2023videodoodles, smith2023method}. It is only recently that automated methods such as Live-Sketch~\cite{gal2024breathing} and FlipSketch~\cite{bandyopadhyay2025flipsketch}, which rely solely on textual guidance, have emerged. 
Given a vector sketch represented by $n$ 2D control points, Live-Sketch uses a set of MLPs to generate a sequence of $f$ sketches, each with $n$ (new) control points. The MLPs are iteratively optimized in a training-data free manner, where the quality of the generated sketches is automatically assessed by a pre-trained text-to-video (T2V) diffusion model \cite{wang2023modelscope} through the Score Distillation Sampling (SDS) technique \cite{poole2022dreamfusion}.
In contrast to Live-Sketch, a given sketch is treated as a raster image in FlipSketch. The raster sketch is first converted to a a noise pattern by DDIM inversion~\cite{song2021denoising} to capture visual appearance. This noise pattern is then fed into a T2V diffusion model~\cite{wang2023modelscope} fine-tuned on training samples synthesized by Live-Sketch to generate an animation. Live-Sketch and FlipSketch demonstrate excellent performance in single-object sketch animation.

Compared to single-object sketch animation, multi-object sketch animation is complex and more challenging, as it not only requires ensuring smooth motion and preserving the visual appearance of each object, but also necessitates considering plausible relationships, interactions and physical constraints among multiple objects. Both Live-Sketch and FlipSketch struggle when transitioning to multi-object scenarios, as shown in \cref{fig:intro}. Live-Sketch does not incorporate object-aware motion modeling, leaving objects' relationships and interactions untouched. The iterative optimization of Live-Sketch process suffers from T2V diffusion models' limitation in handling complex motions among objects, a challenge that has been extensively studied in recent works~\cite{lin2023videodirectorgpt, fei2024dysen, tian2024videotetris, sun2024t2v}. As for FlipSketch, in addition to lacking object-aware motion modeling, the noise pattern extracted through DDIM inversion is insufficient to fully capture the appearance of multi-object sketches. More importantly, since the fine-tuning samples are synthesized from Live-Sketch, which rarely include multi-object scenarios or exhibit poor quality for such cases, severely limiting the multi-object animation performance. 

By analyzing their failures, we summarize that the transition from single-object to multi-object sketch animation presents two challenges: (1) \textbf{object-aware motion modeling}: relative motions, interactions, and physical constraints among objects should be fully considered during motion modeling. (2) \textbf{complex motion optimization}: complex motions of multiple objects should be effectively guided during the optimization process. An effective multi-object sketch animation method should fully address these two challenges.

Since there are no multi-object sketch animation datasets for training, we propose \textbf{MoSketch} for multi-object sketch animation based on iterative optimization through SDS and thus animating a multi-object sketch in a training-data free manner. Following Live-Sketch, we use a vector representation of sketches. We propose four modules: LLM-based scene decomposition, LLM-based motion planning, multi-grained motion refinement and compositional SDS, to tackle the two challenges in a divide-and-conquer strategy. (1) The LLM-based scene decomposition is employed to identify objects, obtain their locations, and decompose complex motions into simpler ones, which serve as the foundation for the other three modules. (2) We employ LLM-based motion planning to generate a motion plan in advance, which defines a coarse object-level motion, utilizing LLM's motion priors about relative motions, interactions and physical constraints. (3) Based on Live-Sketch, we propose multi-grained motion refinement to refine the coarse object motion. (4) During iterative optimization through SDS, we additionally use a compositional SDS to guide the simpler motions modeling sequentially, which can be effectively featured by T2V diffusion models. We briefly compare MoSketch with Live-Sketch and FlipSketch in \cref{tab:intro}. Our key contribution can be summarized as follows:
\begin{itemize}
    \item We propose MoSketch for multi-object sketch animation based on iterative optimization through SDS and thus animating a multi-object sketch in a training-data free manner.
    \item We propose four modules to tackle with the two challenges in multi-object sketch animation in a divide-and-conquer strategy.
    \item Extensive qualitative and quantitative experiments demonstrate the superiority of our method over existing sketch animation approaches.
\end{itemize}

\section{Related Works}
\subsection{Sketch Animation}
Sketch animation aims to convert a static sketch into a dynamic one, which has been extensively studied within the field of computer graphics~\cite{davis2008k, yin2010sketch, gal2024breathing}. Earlier works focus on developing tools
for human-driven creation~\cite{sohn2010sketch, kazi2014draco, ying2024wristsketcher} or incorporating additional inputs such as videos~\cite{suzuki2020realitysketch, yu2023videodoodles, rai2024sketchanim} and skeletons~\cite{dvorovznak2018toonsynth, smith2023method} as references to guide the animation process. 
All these methods need manual intervention. 

Gal \etal \cite{gal2024breathing} propose a text-based sketch animation method Live-Sketch, which does not require any other training data. Instead, it utilizes Score Distillation Sampling (SDS)~\cite{poole2022dreamfusion} to iteratively optimize the animation process by leveraging a pre-defined T2V diffusion model. Some follow-ups~\cite{yang2024sketchanimator, rai2024enhancing} attempt to improve the optimization process of Live-Sketch. By contrast, FlipSketch by Bandyopadhyay \etal \cite{bandyopadhyay2025flipsketch} takes a training-based approach, where a given sketch is first converted to a noise pattern by DDIM inversion. The noise pattern then goes through a T2V diffusion model, fine-tuned on training samples synthesized by Live-Sketch, to generate an animation for the given sketch. Although performing excellently in single-object sketch animation, these methods fail in the multi-object scenario due to the lack of object-aware motion modeling and complex motion optimization. Inheriting the training-data free merit of Live-Sketch, the proposed MoSketch is meant for multi-object sketch animation.


\subsection{Text-guided Image-to-Video Generation}
Close to sketch animation is text-guided image-to-video (I2V) generation~\cite{wang2024videocomposer, hu2024animate, yang2024cogvideox, tang2024any} , aiming to synthesize a video with contents and motions guided from a image and a text description. 
VideoCrafter~\cite{chen2023videocrafter1} is the first open-source foundational model capable of I2V while maintaining content preservation constraints. 
I2VGen-XL~\cite{zhang2023i2vgen} proposes a cascaded framework that first generates low-resolution content consistent video and then performs high-resolution detail refinement.
CogVideoX~\cite{yang2024cogvideox} is based on DiT~\cite{peebles2023scalable}, jointly learning text and visual tokens to perform I2V.
DynamiCrafter~\cite{xing2024dynamicrafter} leverages pre-trained T2V priors with the proposed dual-stream image injection mechanism and the dedicated training paradigm to achieve I2V. 
Although these methods perform well on pixel-domain I2V tasks, they struggle in sketch animation due to the domain gap between sketches and normal images.

\subsection{LLM-assisted Compositional Generation}
Compositional generation requires modeling the relative relationships and interactions between multiple objects, which generative models often struggle to learn effectively~\cite{huang2023t2i, lian2024llm, yan2024dreamdissector}. By contrast, LLMs possess extensive prior knowledge, serving as a valuable complement to address these limitations. 
LLMs have exhibited a strong planning capability in varied text-to-X tasks. In text-to-image~\cite{qu2023layoutllm, qin2024diffusiongpt} and text-to-3D~\cite{zhou2024gala3d, gao2024graphdreamer} compositional generation, LLMs are used for static layout generation of objects, while in T2V~\cite{lian2024llm, tian2024videotetris} and text-to-4D~\cite{xu2024comp4d, zeng2024trans4d}, they are used for dynamic trajectory planning. Furthermore, LLMs have demonstrated scene decomposition capabilities, decomposing complex multi-object task into multiple single- or few-object ones~\cite{lin2023videodirectorgpt, gao2024graphdreamer}. This decomposition enables generative models to tackle the problem in a divide-and-conquer manner.
Inspired by these works, we employ LLM for scene decomposition and motion planning, 
two crucial subtasks for  
multi-object sketch animation.

\section{Proposed Method} \label{sec:method}
Following Live-Sketch, our method uses a vector representation of sketches and achieves multi-object sketch animation based on iterative optimization through SDS. We first introduce the preliminaries of vector sketch representation, followed by a brief description of Live-Sketch. 

\begin{figure*}[htbp]
  \centering
  \includegraphics[width=\textwidth]{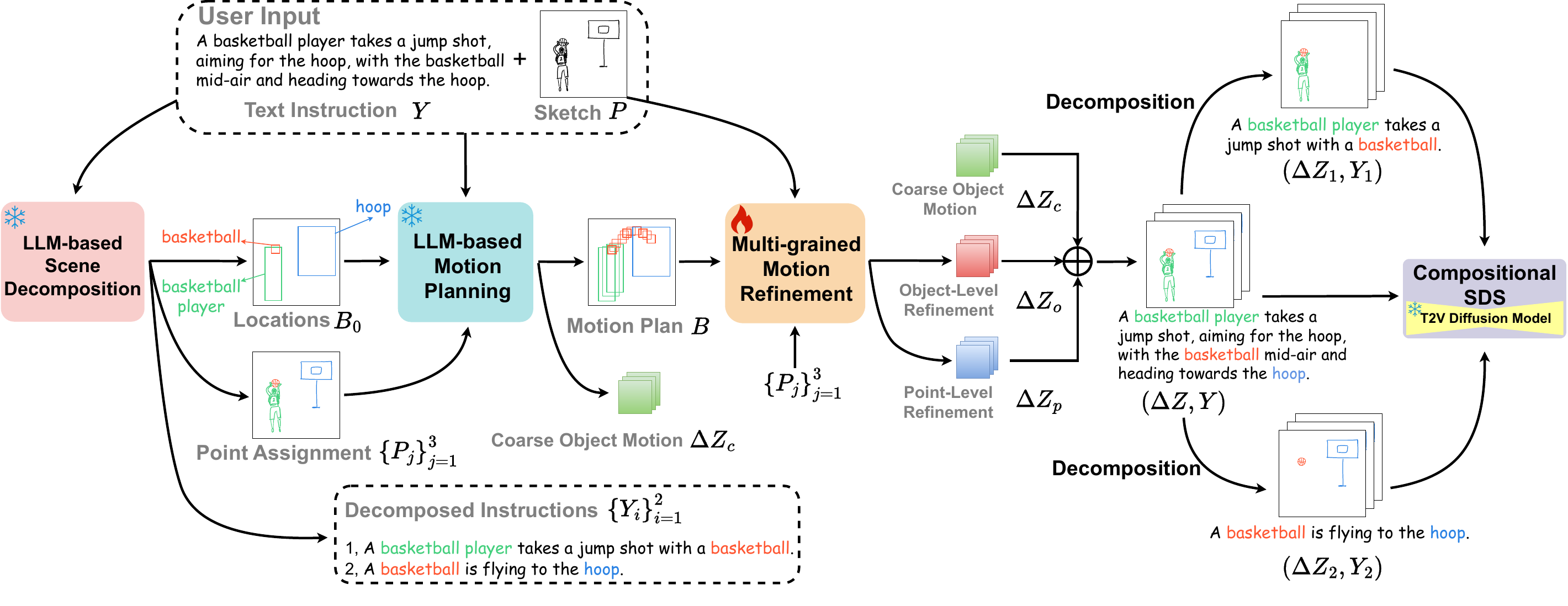}
  \caption{\textbf{Diagram of our proposed MoSketch for multi-object sketch animation.} Four modules: LLM-based scene decomposition, LLM-based motion planning, multi-grained motion refinement and compositional SDS, are proposed to tackle the two challenges of multi-object sketch animation in a divide-and-conquer strategy. There are $m=3$ objects and $r=2$ decomposed instructions in this example.
  } \label{fig:main}
\end{figure*}

\subsection{Preliminaries}

\textbf{Vector Sketch Representation}. 
A vector sketch is composed of strokes, where each stroke is a cubic Bézier curve controlled by four points. A vector sketch can be parameterized by all control points' 2D coordinates. Given a vector sketch $P \in \mathbb{R}^{n \times 2}$ parameterized by $n$ 2D control points and a text instruction $Y$ describing the desired motion, vector sketch animation requires a model to generate a short video consisted of a sequence of vector sketches, formulated by movements of all control points $\Delta Z \in \mathbb{R}^{n \times f \times 2}$, where $f$ is the number of steps, equivalent to the number of frames. This process can be formulated at a high-level as follows:
 \begin{equation}
     \Delta Z \leftarrow \text{Model}(P, Y).
 \end{equation}

\textbf{Live-Sketch in a Nutshell}. 
Live-Sketch designs a simple generative model, separating the sketch animation target $\Delta Z$ into sketch-level motion $\Delta Z_s$ and point-level motion $\Delta Z_p$.
The sketch-level motion $\Delta Z_s$ is the result of holistic transformations (translation, scaling, shearing, and rotation) of the whole sketch, while the point-level motion $\Delta Z_p$ is the translation of control points, focusing on the internal motion of the sketch.

Specially, the vector sketch $P \in \mathbb{R}^{n \times 2}$ is fed into a MLP and get a hidden representation, which is then separated into a sketch embedding $\hat{B} \in \mathbb{R}^{1 \times d}$ and a  point embedding $\hat{P} \in \mathbb{R}^{n \times d}$, where $d$ denotes the hidden dimension. The sketch embedding $\hat{B} \in \mathbb{R}^{1 \times d}$ is passed to a MLP to predict parameters $\widetilde{B} \in \mathbb{R}^{f \times 7}$ of holistic transformations in all frames (seven parameters in a frame: two for translation, two for scaling, two for shearing and one for rotation). The holistic transformations are applied to all control points $P$, yielding a sketch-level motion $\Delta Z_s \in \mathbb{R}^{n \times f \times 2}$. The point embedding directly predicts all control points' translations $\Delta Z_p \in \mathbb{R}^{n \times f \times 2}$ through a MLP. The sketch-level motion $\Delta Z_s$ and point-level motion $\Delta Z_p$ are added to get the sketch animation $\Delta Z$. This process is formulated as:
\begin{equation}
\resizebox{0.6\hsize}{!}{$
\left\{
\begin{array}{lll}
\hat{B}, \hat{P}& \leftarrow &  \text{MLP}(P),\\
\widetilde{B} & \leftarrow & \text{MLP}(\hat{B}), \\
\Delta Z_s & \leftarrow &  \text{transformation}(\widetilde{B},P),\\
\Delta Z_p & \leftarrow &  \text{MLP}(\hat{P}),\\
\Delta Z & \leftarrow &  \Delta Z_s + \Delta Z_p.\\
\end{array}
\right. $}
\end{equation}

Live-Sketch utilizes Score Distillation Sampling (SDS)~\cite{poole2022dreamfusion} to leverage a pre-trained T2V diffusion model~\cite{wang2023modelscope} to guide the animation process. SDS is a method that using a pre-trained pixel-aware diffusion model to guide other non-pixel generation process in an iterative optimization-based manner, without any other data for training. We denote the SDS loss in Live-Sketch as:
\begin{equation}
     \mathcal{L}_{SDS} \leftarrow \text{SDS}(\Delta Z, Y).
\end{equation}
By iteratively minimizing $\mathcal{L}_{SDS}$, the animation result $\Delta Z$ will progressively align with the text instruction $Y$. 

\subsection{Multi-object Sketch Animation}
Based on Live-Sketch, we propose MoSketch for multi-object sketch animation, as shown in \cref{fig:main}. We propose four modules: LLM-based scene decomposition, LLM-based motion planning, multi-grained motion refinement and compositional SDS, to tackle the two challenges described in Introduction in a divide-and-conquer strategy. The LLM-based scene decomposition is the foundation of other three modules, which is employed to identify objects, obtain their locations, and decompose complex motions into simpler components. Based on it, the LLM-based motion planning and the multi-grained motion refinement achieve the object-aware motion modeling considering of relative motions, interactions and physical constraints among objects. The compositional SDS ensures that the complex motions of multiple objects are effectively guided during the iterative optimization. We will describe these four modules as follows. 

\begin{figure}[htbp]
    \centering
    \begin{subfigure}[t]{\linewidth}
        \centering
        \includegraphics[width=\linewidth]{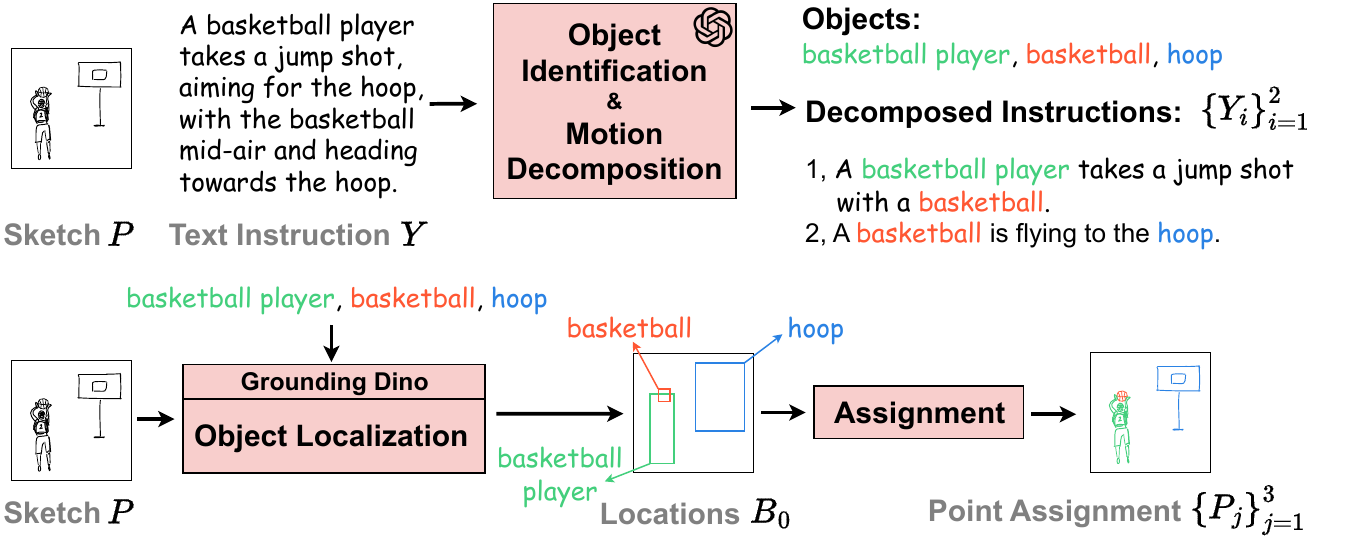}
        \caption{LLM-based scene decomposition}
        \label{fig:scene}
    \end{subfigure}
    \begin{subfigure}[t]{\linewidth}
        \centering
        \includegraphics[width=\linewidth]{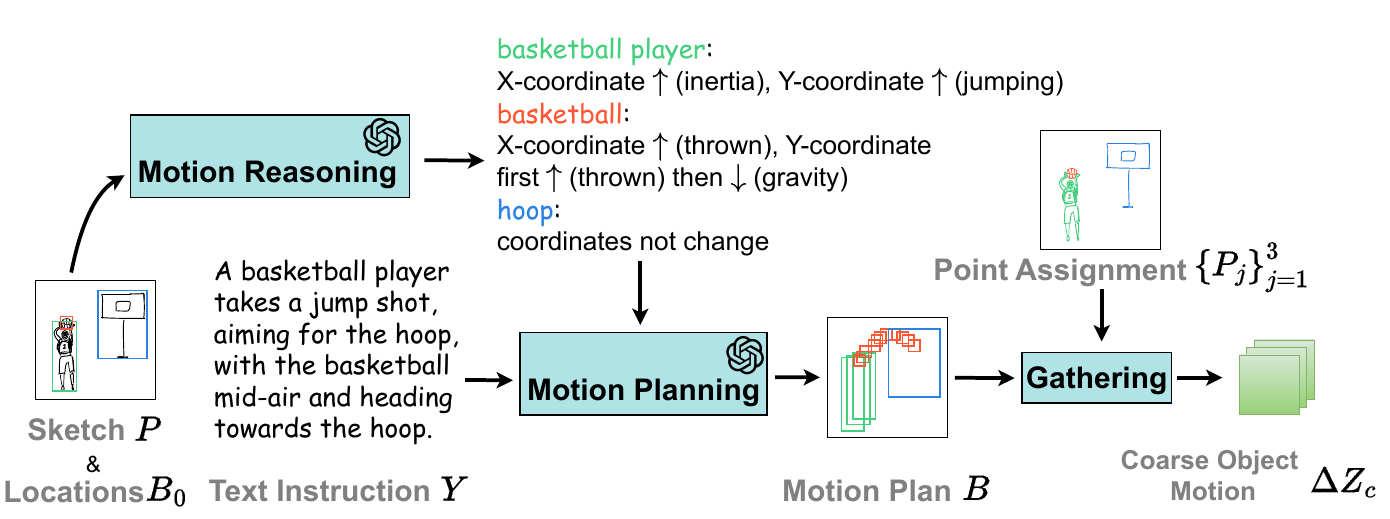}
        \caption{LLM-based motion planning}
        \label{fig:plan}
    \end{subfigure}
    \begin{subfigure}[t]{\linewidth}
        \centering
        \includegraphics[width=\linewidth]{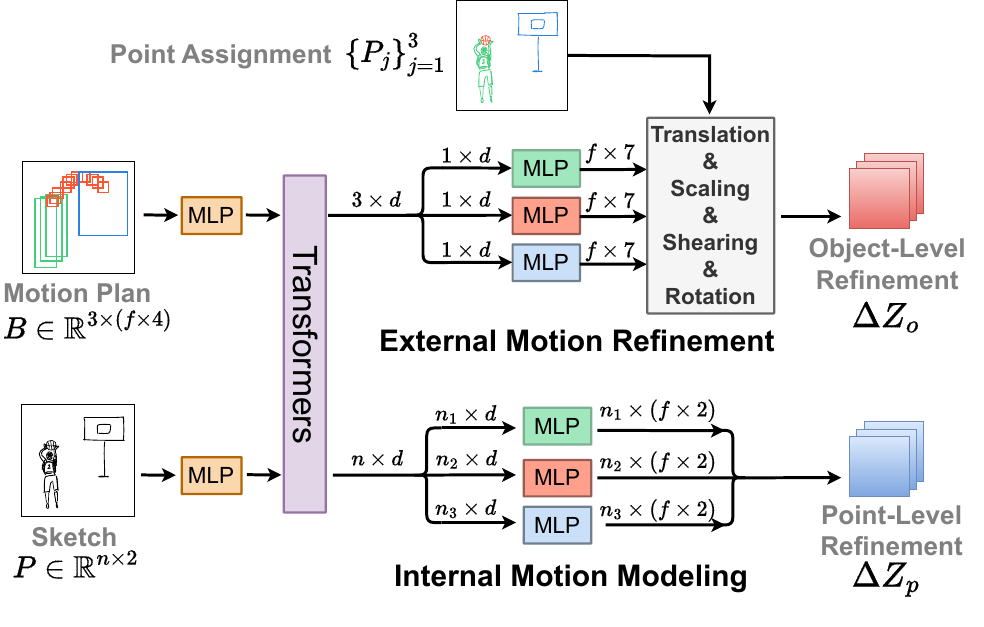}
        \caption{Multi-grained motion refinement}
        \label{fig:network}
    \end{subfigure}
    \caption{\textbf{Illustration of the three modules for object-aware motion modeling.} (a) \textbf{LLM-based scene decomposition:} used to identify objects, obtain their locations, and decompose complex motions into simpler ones. (b) \textbf{LLM-based motion plan:} defining a coarse object external motion for a multi-object sketch. (c) \textbf{Multi-grained motion refinement:} generating a object-level refinement and a point-level refinement for external motion refinement and internal motion modeling of objects respectively. $n_1,n_2,n_3$ are the number of control points in the three objects.}
    \label{fig:modules}
\end{figure}

\subsubsection{LLM-based Scene Decomposition} \label{ssec:scene}
The scene decomposition is employed to identify objects, obtain their locations, and decompose complex motions into simpler components, which serves as the foundation for the other three modules. We employ GPT-4 for scene decomposition, which is successfully applied in compositional generation works~\cite{lin2023videodirectorgpt, su2023motionzero, xu2024comp4d}, as shown in \cref{fig:scene}. Given a sketch $P$ and a text instruction $Y$, we first ask GPT-4 to identify the objects requiring motion planning and decompose the complex motion described in $Y$, resulting in $m$ identified objects and $r$ simple motions. The $m$ identified objects should be independent of each other, with no hierarchical or synonymous relationships. The decomposed $r$ simpler motions are described in $r$ short text instructions $\{Y_i\}_{i=1}^r$, and each should involve one or few identified objects~\cite{lin2023videodirectorgpt, xu2024comp4d}. In our method, both $m$ and $r$ are not fixed, but $m$ should be no more than 7 and $r$ should be no more than 5. We employ a powerful open-world object detection model Grounding DINO~\cite{liu2024grounding} to get objects' bounding boxes $B_0 \in \mathbb{R}^{m \times 4}$. Note that sketch $P$ is parameterized by the $n$ control points, we assign each point to an object based on the distance between the center point of the stroke it controls and the bounding boxes of all objects. The point is assigned to the object whose bounding box is closest to the controlling stroke's center. The assignment result is denoted as $\{P_j\}_{j=1}^m$, where $P_j$ are control points belonging to the $j$-th object. The LLM-based scene decomposition can be formulated briefly as:
\begin{equation}
\resizebox{0.83\hsize}{!}{$
\left\{
\begin{array}{lll}
\text{objects}, \{Y_i\}_{i=1}^r & \leftarrow &  \text{GPT4}(P, Y),\\
B_0 & \leftarrow & \text{grounding}(P, \text{objects}), \\
\{P_j\}_{j=1}^m & \leftarrow & \text{assign}(B_0, P). \\
\end{array}
\right. $}
\end{equation}

\subsubsection{LLM-based Motion Planning} \label{ssec:plan}
Object-aware motion modeling should consider relative motions, interactions, and particularly physical constraints among objects, which are important yet difficult to model with conventional networks like Live-Sketch. Recent works~\cite{lv2024gpt4motion, zhu2024compositional} reveal that GPT-4 possesses prior knowledge of multi-object motions in the real world and can roughly plan the external motion of objects. Thus we employ GPT-4 for motion planning, defining a coarse object-level motion for object-aware motion modeling, as shown in \cref{fig:plan}. Given a sketch $P$, a text instruction $Y$, objects' initial location $B_0$, GPT-4 generates a coarse motion plan which is the bounding boxes of objects in $f$ frames, denoted as $B \in \mathbb{R}^{m \times f \times 4}$. To ensure that GPT-4 fully considers object interactions, relative motions and physical constraints such as inertia and gravity during motion planning, we follow \cite{lian2024llm} by incorporating a reasoning step before GPT-4 generates its response. With the motion plan $B$ and point assignment $\{P_j\}_{j=1}^m$, we gather a coarse object motion $\Delta Z_c \in \mathbb{R}^{n \times f \times 2}$. This process is formulated as:
\begin{equation}
\resizebox{0.65\hsize}{!}{$
\left\{
\begin{array}{lll}
B & \leftarrow &  \text{GPT4}(P, Y + B_0),\\
\Delta Z_c & \leftarrow & \text{gather}(\{P_j\}_{j=1}^m, B). \\
\end{array}
\right. $}
\end{equation}

\begin{figure*}[htbp]
  \centering
  \includegraphics[width=\textwidth]{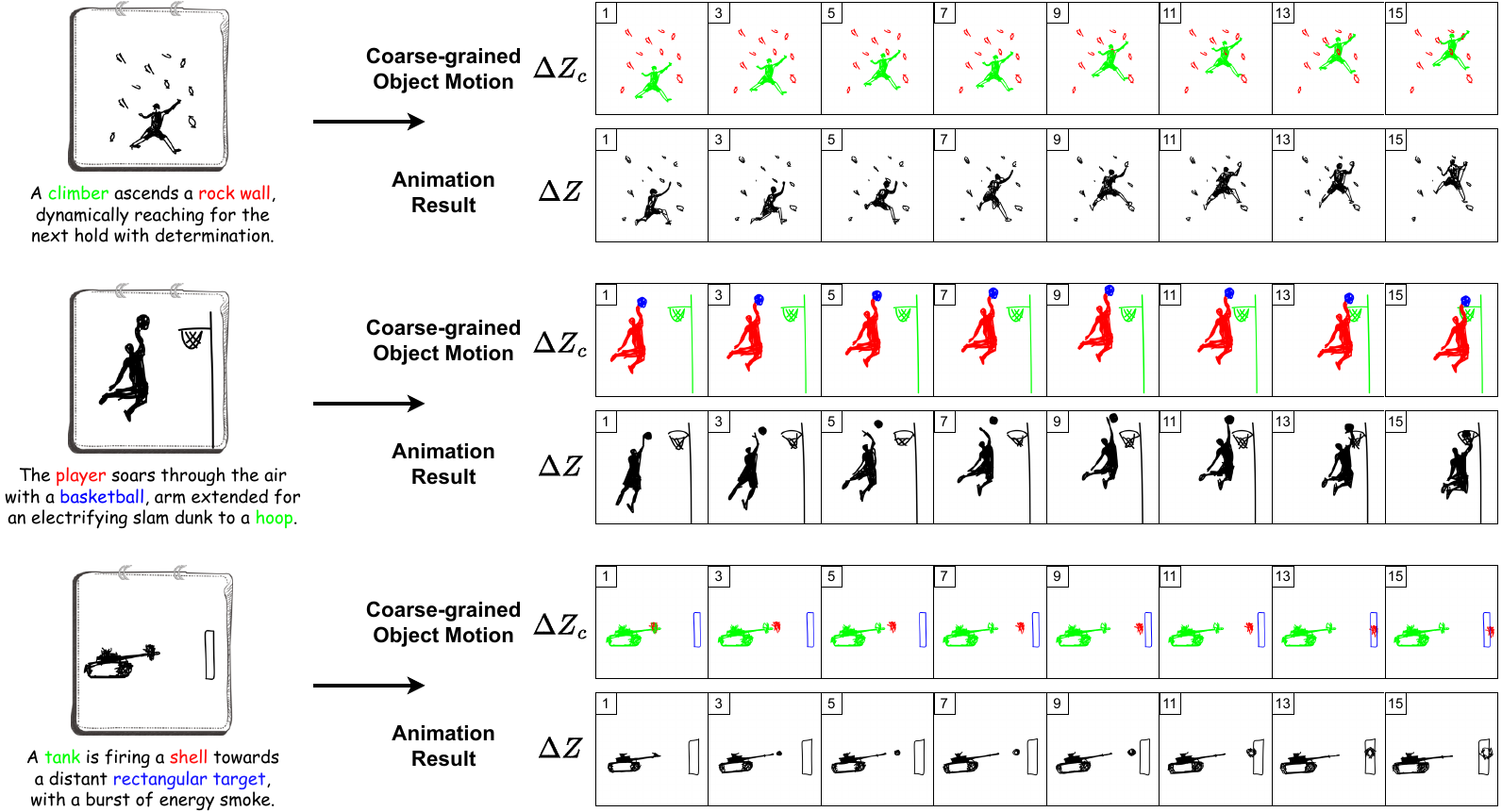}
  \caption{\textbf{Qualitative results}. We exhibit the multi-object sketch animation results $\Delta Z$ and coarse-grained object motion $\Delta Z_c$ predefined by LLM-based motion planning. More results are provided in the supplementary. Best view digitally.} \label{fig:result}
\end{figure*}

\subsubsection{Multi-grained Motion Refinement} \label{ssec:fine}
While GPT-4 provides a motion plan, a generative network is still required to refine the external motion and model the internal motion of objects. Based on Live-Sketch, we propose multi-grained motion refinement, as shown in \cref{fig:network}. We replace the sketch-level motion $\Delta Z_s$ in Live-Sketch to the object-level motion $\Delta Z_o$, and turn the point-level motion $\Delta Z_p$ in Live-Sketch to the object-aware one. We regard $\Delta Z_o$ and $\Delta Z_p$ to the refinement of external motion and the modeling of internal motion for all objects, respectively. 

Specially, besides $P$ as the input of all points, we add the motion plan $B \in \mathbb{R}^{m \times (f \times 4)}$ as the input of $m$ objects. Each passes through a MLP to get a hidden representation. The two hidden representations are then concatenated and fed into Transformers with self attention units and positional encodings \cite{liu2024transformer} specially designed for vector sketches to feature relationships and interactions between objects, yielding object embedding $\hat{B} \in \mathbb{R}^{m \times d}$ and point embedding $\hat{P} \in \mathbb{R}^{n \times d}$: 
\begin{equation}
\hat{B},\hat{P} \leftarrow \text{Transformers}(\text{MLP}(B),\text{MLP}(P)).\\
\end{equation}

Similar to sketch embedding for holistic transformations of the whole sketch in Live-Sketch, object embeddings $\hat{B}$ are used for holistic transformations of objects. Each object embedding $\hat{B}_j$ is passed through a dedicated MLP to predict the seven transformation parameters $\widetilde{B}_j \in \mathbb{R}^{f \times 7}$ and the transformations are applied to relative control points $P_j$. We gather all transformation results as the object-level refinement $\Delta Z_o \in \mathbb{R}^{n \times f \times 2}$. Like Live-Sketch, point embedding $\hat{P}$ directly predicts points' translations $\Delta Z_p \in \mathbb{R}^{n \times f \times 2}$, with different MLPs for points in different objects. Finally, the coarse object motion $\Delta Z_c$, object-level refinement $\Delta Z_o$ and point-level refinement $\Delta Z_p$ are added to generate sketch animation $\Delta Z$. This process is denoted as: 
\begin{equation}
\resizebox{0.9\hsize}{!}{$
\left\{
\begin{array}{lll}
\{\hat{B}_j\}_{j=1}^m, \{\hat{P}_j\}_{j=1}^m & \leftarrow &  \hat{B}, \hat{P},\\
\{\widetilde{B}_j\}_{j=1}^m & \leftarrow &  \{\text{MLP}_j(\hat{B}_j)\}_{j=1}^m,\\
\Delta Z_o & \leftarrow &  \{\text{transformation}(\widetilde{B}_j,P_j)\}_{j=1}^m,\\
\Delta Z_p & \leftarrow &  \{\text{MLP}_j(\hat{P}_j)\}_{j=1}^m,\\
\Delta Z & \leftarrow &  \Delta Z_c + \Delta Z_o + \Delta Z_p.\\
\end{array}
\right. $}
\end{equation}

\subsubsection{Compositional SDS} \label{ssec:csds}
Following Live-Sketch, we use SDS to leverage a pre-trained T2V diffusion model~\cite{wang2023modelscope} to iteratively guide the animation, without any other data for training. Compared to normal generation, compositional generation better understands inter-object relationships. To ensure that the complex motions of multiple objects are effectively guided, inspired by compositional generation works~\cite{zhou2024gala3d, gao2024graphdreamer, xu2024comp4d}, we use compositional SDS in addition to the original one in Live-Sketch during the T2V diffusion model guidance. 
Specially, for each decomposed instruction $Y_i$, per $\Delta Z$ frame expressed by a set of control points $P$, we extract a point subset $P^{'}_i$ that exclusively cover all objects specified by $Y_i$. Putting $P^{'}_i$ from all frames together, we obtain the sub-video $\Delta Z_i$. A SDS loss $\mathcal{L}_{SDS-i}$ is calculated upon $\Delta Z_i$ and $Y_i$, guiding the simpler motion in $Y_i$, which can be effectively featured by the T2V diffusion model. All SDS loss of simple motions $\{Y_i\}_{i=1}^r$ are added with the original SDS loss $\mathcal{L}_{SDS}$. The guidance process is formulated as:
\begin{equation}
\resizebox{0.85\hsize}{!}{$
\left\{
\begin{array}{lll}
\{P^{'}_i\}_{i=1}^r & \leftarrow &  \text{extract}(P, \{Y_i\}_{i=1}^r),\\
\{\Delta Z_i\}_{i=1}^r & \leftarrow &  \text{decompose}(\Delta Z, \{P^{'}_i\}_{i=1}^r),\\
\{\mathcal{L}_{SDS-i}\}_{i=1}^r & \leftarrow &  \{\text{SDS}(\Delta Z_i, Y_i)\}_{i=1}^r,\\
\mathcal{L}_{SDS} & \leftarrow &  \text{SDS}(\Delta Z, Y),\\
\mathcal{L}_{CSDS} & \leftarrow & \mathcal{L}_{SDS} + \sum_{i=1}^r \mathcal{L}_{SDS-i}. \\
\end{array}
\right. $} 
\label{eq:sds}
\end{equation}

\section{Evaluation}

\subsection{Experimental Setup}

\textbf{Testing Data Creation}.
We create 60 multi-object sketches to test multi-object sketch animation. First, we random select pixel-based images with at least two objects from three categories: human, animal and object following~\cite{gal2024breathing}. Then we use CLIPasso~\cite{vinker2022clipasso} to convert these images to vector sketches. Finally, for each multi-object sketch, we employ GPT-4 to generate a text description which implicitly suggests possible motions. These 60 sketches encompass various real-life scenarios such as sports, dining, transportation and work, as provided in the supplementary.

\textbf{Baselines}. We compare our method with two 
 text-guided sketch animation methods:  \\
 $\bullet$ Live-Sketch, CVPR24 \cite{gal2024breathing}, which uses a vector representation and sketches and employ SDS to leverage a pre-trained T2V diffusion model for animation without any other data for fine-tuning. \\ 
 $\bullet$ FlipSketch, CVPR25 \cite{bandyopadhyay2025flipsketch}, which applies DDIM inversion to the given raster sketch and perform sketch animation by a fine-tuned T2V diffusion model. 
 
 Viewing text-guided sketch animation as a special case of text-guided image-to-video (I2V) generation, we further compare with two I2V methods: \\
 $\bullet$ CogVideoX, arxiv24 \cite{yang2024cogvideox}, a DiT-based method that jointly learning text and visual tokens. \\
 $\bullet$ DynamiCrafter, ECCV24 \cite{xing2024dynamicrafter}, leveraging pre-trained T2V priors with a dual-stream image injection mechanism and dedicated training paradigm to achieve I2V generation.

\textbf{Details of Implementation}.
The parameters in the multi-grained motion refinement should be  optimized.
The hidden dimension $d$ of both object and point embedding is set to 128. The number of frames $f$ is fixed at 16. The Transformers consists of 2 layers. For optimization, we use the Adam optimizer with an initial learning rate of 5e-3 and a weight decay of 1e-2. The multi-grained motion refinement is iterated for 500 steps, requiring approximately one hour on a single RTX 3090 Ti GPU.

\textbf{Evaluation Criteria}. 
We follow Live-Sketch~\cite{gal2024breathing} and FlipSketch~\cite{bandyopadhyay2025flipsketch}, evaluating the model’s ability to generate videos that align with the text instruction (``Text-to-Video Alignment"),  as well as the ability to preserve structural characteristics (``Sketch-to-Video Alignment"). We employ ``Overall Consistency" and ``I2V Subject" introduced in a comprehensive video generation benchmark VBench~\cite{huang2024vbench, huang2024vbench++} as the metric for ``Text-to-Video Alignment" and ``Sketch-to-Video Alignment" respectively. Note that the ``Sketch-to-Video Alignment" metric exhibits an inherent bias: it yields inflated scores when the target video contains objects with minimal shape variation relative to the input sketch. Additionally, we use ``Motion Smoothness" and ``Dynamic Degree" from VBench to evaluate the smoothness and dynamics of generated videos.

\begin{figure}
  \centering
  \includegraphics[width=\linewidth]{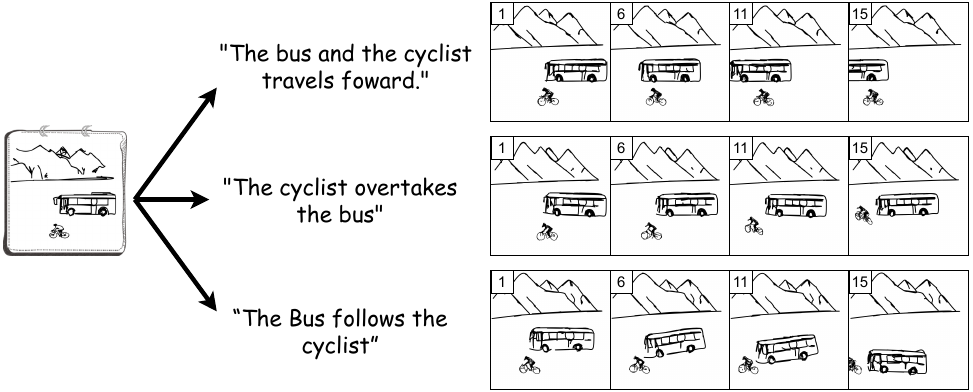}
  \caption{\textbf{Different animation results according to different text instructions}. Best view digitally.} \label{fig:text}
\end{figure}

\textbf{Qualitative Results}. 
\cref{fig:result} shows the multi-object sketch animation results of MoSketch. With the LLM-based motion planning and the multi-grained motion refinement, MoSketch achieves vivid and realistic animations of complex scenarios, including rock climbing, basketball playing, and tank target shooting. The LLM-based motion plans are highly plausible, while the multi-grained motion refinement further refines the animation (\eg \emph{the basketball finally enters the net, and the shell explodes when it collides with the target}). We observe that even when the object localization or point assignment during LLM-based scene description contains minor inaccuracies (\eg \emph{the smoke is not grounded and the shell's point assignment is inaccurate}), the final results remain robust and visually compelling. \cref{fig:text} demonstrates that, given a scene sketch, MoSketch generates diverse animation results by considering different object interactions or relative relationships based on varying text instructions. This capability greatly improves the diversity and flexibility of our approach. Additional evaluations on freehand multi-object sketches and single-object sketches are provided in the supplementary.

\begin{figure}
  \centering
  \includegraphics[width=\linewidth]{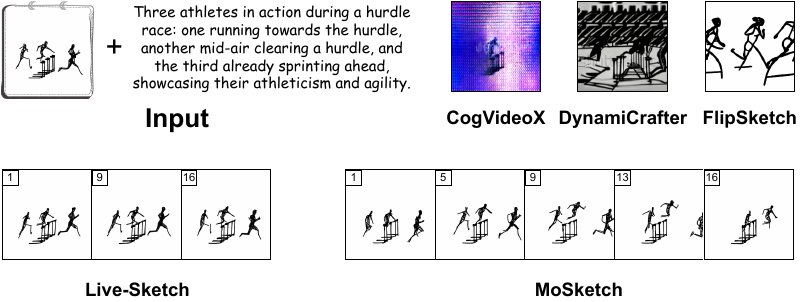}
  \caption{\textbf{Qualitative comparisons for multi-object sketch animation}. Best view digitally.} \label{fig:quality}
\end{figure}

\begin{table}
\centering
\resizebox{\linewidth}{!}{
\begin{tabular}{lcccc}
\toprule
\multirow{2}*{\textbf{Method}} & Text-to-Video & Sketch-to-Video & Motion & Dynamic \\
                               &  Alignment & Alignment & Smoothing & Degree \\

\midrule
CogVideoX                        & 0.141 & 0.610 & 0.747 & -  \\
DynamiCrafter                        & 0.184 & 0.771 & 0.868 & -  \\
FlipSketch                            & 0.199 & 0.704  & 0.839  & -  \\
Live-Sketch                             & 0.207 & 0.897 & 0.956 & 0.266  \\
\midrule
MoSketch                               & \textbf{0.218} & \textbf{0.914} & \textbf{0.977} & \textbf{0.283} \\
\bottomrule
\end{tabular}}
\caption{\textbf{Quantitative comparisons for multi-object sketch animation}. Due to the failure to preserve visual appearance in CogVideoX, DynamiCrafter, and FlipSketch, the ``Dynamic Degree" metric in these methods lacks meaningful interpretation.}
\label{tab:quantity}
\end{table}

The result of qualitative comparison is shown in \cref{fig:quality}. Due to the domain gap between natural images and sketches, and the lack of specialized training on sketch data, the results generated by the I2V methods CogVideoX and DynamiCrafter fail to preserve visual appearance of the input sketch, leading to chaos. FlipSketch generates animation results in the sketch domain due to fine-tuning on sketch data, but fails to preserve visual appearance due to the raster representation of sketches. Live-Sketch preserves visual appearance due to the vector representation of sketches but struggles to model complex motion in multi-object animation. Our proposed MoSketch designs three effective modules to handle complex motion modeling, leading to vivid and realistic multi-object sketch animation.

\textbf{Quantitative Comparison}. 
\cref{tab:quantity} shows the quantitative comparisons of MoSketch and baselines. In ``Text-to-Video Alignment" and ``Sketch-to-Video Alignment", MoSketch achieves superior performance. Due to inability to follow the complex instructions and preserve visual appearance, CogVideoX, DynamiCrafter and FlipSketch get low scores in ``Text-to-Video Alignment" and ``Sketch-to-Video Alignment". Live-Sketch preserves visual appearance because of the vector representation of sketches, but gets lower ``Text-to-Video Alignment" score than MoSketch due to the inability to generate complex motions required in text instructions. For the evaluation of video quality, MoSketch achieves the smoothest motion. Compared to Live-Sketch, MoSketch's animations are more dynamic.

\subsection{Ablation Study of MoSketch}
\begin{figure}
  \centering
  \includegraphics[width=\linewidth]{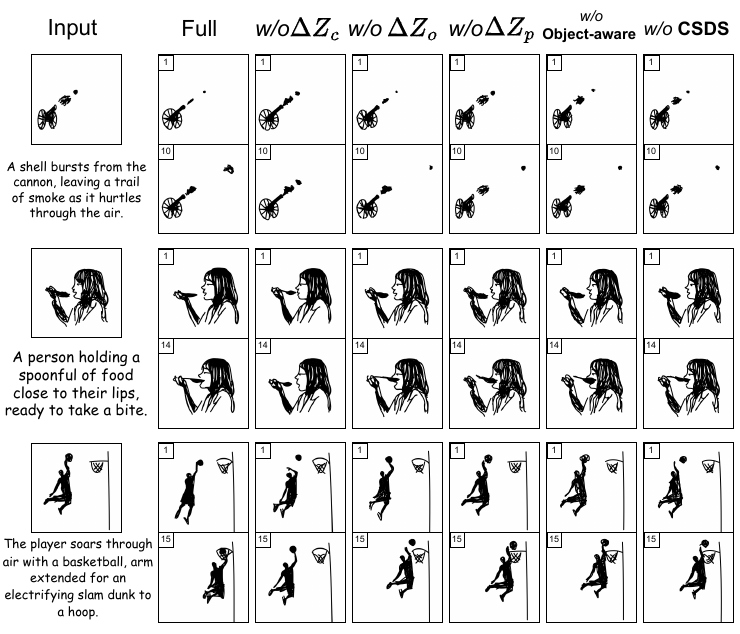}
  \caption{\textbf{Qualitative ablation of MoSketch}. Best view digitally.}  \label{fig:ablation}
\end{figure}

We analyze the effectiveness of LLM-based motion plan, multi-grained motion refinement and compositional SDS with several ablation studies. The qualitative and quantitative ablation study results are shown in \cref{fig:ablation} and \cref{tab:ablation}. 

\begin{table}
\centering
\resizebox{\linewidth}{!}{
\begin{tabular}{llcccc}
\toprule
\multirow{2}*{\textbf{\#}} & \multirow{2}*{\textbf{setup}} & Text-to-Video & Sketch-to-Video & Motion & Dynamic \\
                     &      &  Alignment & Alignment & Smoothing & Degree \\
\midrule
0 & Full                               & \textbf{0.218} & 0.914 & \textbf{0.977} & \textbf{0.283} \\
\midrule
1 & \textit{w/o} $\Delta Z_c$       & 0.212 & 0.955 & 0.959 & 0.083 \\
2 & \textit{w/o} $\Delta Z_o$       & 0.212 & 0.909 & 0.964 & 0.266 \\
3 & \textit{w/o} $\Delta Z_p$       & 0.203 & \textbf{0.971} & 0.971 & 0.200 \\
4 & \textit{w/o} Object-aware    & 0.205 & 0.932 & 0.968 & 0.266 \\
5 & \textit{w/o} CSDS                   & 0.207 & 0.911 & 0.966 & 0.267 \\
\bottomrule
\end{tabular}}
\caption{\textbf{Quantitative ablation of MoSketch}. Note that Setup\#1, Setup\#3 and Setup\#4 get inflated scores in ``Sketch-to-Video Alignment" due to its inherent bias.}
\label{tab:ablation}
\end{table}

\textbf{The Need of Motion Planning}. We eliminate the coarse object motion $\Delta Z_c$ defined by the generated motion plan (Setup\#1), leading to nearly static external movements (\eg \emph{stalled shell launch}).

\textbf{The Necessity of Multi-grained Motion Refinement}. We separately remove the object-level refinement $\Delta Z_o$ (Setup\#2) and the point-level refinement $\Delta Z_p$ (Setup\#3). Without $\Delta Z_o$, the object external motion could not be refined (\eg \emph{failed basketball-hoop entries}). The absence of $\Delta Z_p$ leads to the lack of internal motion within objects. 

\textbf{The Necessity of Object-aware Network}. We replace our object-aware multi-grained motion refinement with the not object-aware network in Live-Sketch (Setup\#4). The lack of object-aware motion modeling in generative network results in semantically implausible object interactions (\eg \emph{misaligned spoon-nose placements}).

\textbf{The Need of Compositional Optimization}. We remove the SDS loss $\{\mathcal{L}_{SDS-i}\}_{i=1}^r$ in \cref{eq:sds} (Setup\#5), only using $\mathcal{L}_{SDS}$ to guide the complex motion modeling. Guided by a T2V diffiusion model that struggles with modeling complex motions of multiple objects, the generated animations naturally lack details.

\begin{figure}[t]
    \centering
    \begin{subfigure}[t]{\linewidth}
        \centering
        \includegraphics[width=\linewidth]{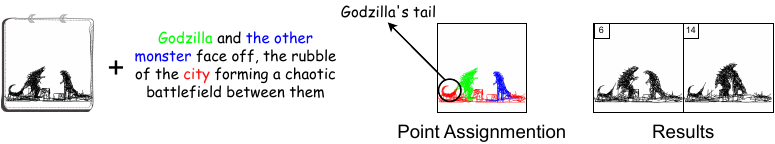}
        \caption{Incorrect point assignment (\emph{of Godzilla's tail})}
        \label{fig:limit1}
    \end{subfigure}
    \begin{subfigure}[t]{\linewidth}
        \centering
        \includegraphics[width=\linewidth]{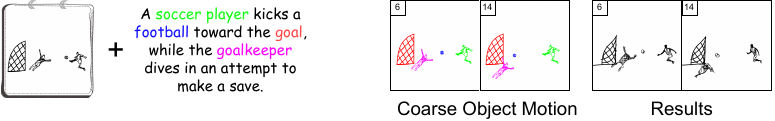}
        \caption{Incorrect motion planning (\emph{goalkeeper shall move towards the ball})}
        \label{fig:limit2}
    \end{subfigure}
    \begin{subfigure}[t]{\linewidth}
        \centering
        \includegraphics[width=\linewidth]{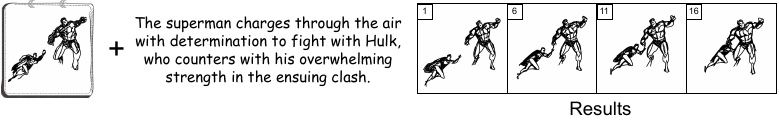}
        \caption{Failed to generate specified motion (\emph{fight})}
        \label{fig:limit3}
    \end{subfigure}
    \caption{\textbf{Failure cases}.}\label{fig:limit}
\end{figure}


\textbf{Limitations}. 
\cref{fig:limit} shows several limitations of our method: (1) While \cref{fig:result} illustrates that MoSketch can tolerate minor point assignment inaccuracies without significantly degrading the final results, a large number of errors can severely impact the output quality (\eg \emph{Godzilla's tail is incorrectly assigned to ``city"}). (2) Coarse object motion derived from a highly incorrect motion plan (\eg \emph{the goalkeeper should move towards the football}) cannot be corrected by the multi-grained motion refinement. (3) Since our animations are guided by a T2V diffusion model~\cite{wang2023modelscope}, which is unaware of specified motion such as \emph{fight}, the relative animation could not be generated successfully.

\section{Conclusions}

We propose MoSketch for multi-object sketch animation based on iterative optimization through SDS and thus animating a multi-object sketch in a training-data free manner. To tackle the two challenges: object-aware motion modeling and complex motion optimization, we propose four modules: LLM-based scene decomposition, LLM-based motion planning, multi-grained motion refinement and compositional SDS. Extensive experiments reveal that our proposed MoSketch achieves the superior performance than the SOTA methods in multi-object sketch animation. The ablation studies demonstrate the effectiveness and necessity of the proposed modules. Other experiments show the flexibility of MoSketch. MoSketch takes a pioneering step towards multi-object sketch animation, opening new avenues for future research and applications.

\section*{Acknowledgments}
This work was supported by NSFC (No. 62172420) and Beijing Natural Science Foundation (No. L254039).

{
    \small
    \bibliographystyle{ieeenat_fullname}
    \bibliography{main}

\begin{thebibliography}{50}
\providecommand{\natexlab}[1]{#1}
\providecommand{\url}[1]{\texttt{#1}}
\expandafter\ifx\csname urlstyle\endcsname\relax
  \providecommand{\doi}[1]{doi: #1}\else
  \providecommand{\doi}{doi: \begingroup \urlstyle{rm}\Url}\fi

\bibitem[Bandyopadhyay and Song(2025)]{bandyopadhyay2025flipsketch}
Hmrishav Bandyopadhyay and Yi-Zhe Song.
\newblock Flipsketch: Flipping static drawings to text-guided sketch animations.
\newblock In \emph{CVPR}, 2025.

\bibitem[Chen et~al.(2023)Chen, Xia, He, Zhang, Cun, Yang, Xing, Liu, Chen, Wang, et~al.]{chen2023videocrafter1}
Haoxin Chen, Menghan Xia, Yingqing He, Yong Zhang, Xiaodong Cun, Shaoshu Yang, Jinbo Xing, Yaofang Liu, Qifeng Chen, Xintao Wang, et~al.
\newblock Videocrafter1: Open diffusion models for high-quality video generation.
\newblock \emph{arXiv preprint arXiv:2310.19512}, 2023.

\bibitem[Chowdhury et~al.(2022)Chowdhury, Sain, Bhunia, Xiang, Gryaditskaya, and Song]{chowdhury2022fs}
Pinaki~Nath Chowdhury, Aneeshan Sain, Ayan~Kumar Bhunia, Tao Xiang, Yulia Gryaditskaya, and Yi-Zhe Song.
\newblock Fs-coco: Towards understanding of freehand sketches of common objects in context.
\newblock In \emph{ECCV}, pages 253--270, 2022.

\bibitem[Davis et~al.(2008)Davis, Colwell, and Landay]{davis2008k}
Richard~C Davis, Brien Colwell, and James~A Landay.
\newblock K-sketch: a'kinetic'sketch pad for novice animators.
\newblock In \emph{CHI}, pages 413--422, 2008.

\bibitem[Dvoro{\v{z}}n{\'a}k et~al.(2018)Dvoro{\v{z}}n{\'a}k, Li, Kim, and S{\`y}kora]{dvorovznak2018toonsynth}
Marek Dvoro{\v{z}}n{\'a}k, Wilmot Li, Vladimir~G Kim, and Daniel S{\`y}kora.
\newblock Toonsynth: example-based synthesis of hand-colored cartoon animations.
\newblock \emph{ACM Transactions on Graphics}, 37\penalty0 (4):\penalty0 1--11, 2018.

\bibitem[Fei et~al.(2024)Fei, Wu, Ji, Zhang, and Chua]{fei2024dysen}
Hao Fei, Shengqiong Wu, Wei Ji, Hanwang Zhang, and Tat-Seng Chua.
\newblock Dysen-vdm: Empowering dynamics-aware text-to-video diffusion with llms.
\newblock In \emph{CVPR}, pages 7641--7653, 2024.

\bibitem[Gal et~al.(2024)Gal, Vinker, Alaluf, Bermano, Cohen-Or, Shamir, and Chechik]{gal2024breathing}
Rinon Gal, Yael Vinker, Yuval Alaluf, Amit Bermano, Daniel Cohen-Or, Ariel Shamir, and Gal Chechik.
\newblock Breathing life into sketches using text-to-video priors.
\newblock In \emph{CVPR}, pages 4325--4336, 2024.

\bibitem[Gao et~al.(2024)Gao, Liu, Chen, Geiger, and Sch{\"o}lkopf]{gao2024graphdreamer}
Gege Gao, Weiyang Liu, Anpei Chen, Andreas Geiger, and Bernhard Sch{\"o}lkopf.
\newblock Graphdreamer: Compositional 3d scene synthesis from scene graphs.
\newblock In \emph{CVPR}, pages 21295--21304, 2024.

\bibitem[Ha and Eck(2018)]{ha2018neural}
David Ha and Douglas Eck.
\newblock A neural representation of sketch drawings.
\newblock In \emph{ICLR}, 2018.

\bibitem[Hu(2024)]{hu2024animate}
Li Hu.
\newblock Animate anyone: Consistent and controllable image-to-video synthesis for character animation.
\newblock In \emph{CVPR}, pages 8153--8163, 2024.

\bibitem[Huang et~al.(2023)Huang, Sun, Xie, Li, and Liu]{huang2023t2i}
Kaiyi Huang, Kaiyue Sun, Enze Xie, Zhenguo Li, and Xihui Liu.
\newblock T2i-compbench: A comprehensive benchmark for open-world compositional text-to-image generation.
\newblock \emph{Advances in Neural Information Processing Systems}, 36:\penalty0 78723--78747, 2023.

\bibitem[Huang et~al.(2024{\natexlab{a}})Huang, He, Yu, Zhang, Si, Jiang, Zhang, Wu, Jin, Chanpaisit, et~al.]{huang2024vbench}
Ziqi Huang, Yinan He, Jiashuo Yu, Fan Zhang, Chenyang Si, Yuming Jiang, Yuanhan Zhang, Tianxing Wu, Qingyang Jin, Nattapol Chanpaisit, et~al.
\newblock Vbench: Comprehensive benchmark suite for video generative models.
\newblock In \emph{CVPR}, pages 21807--21818, 2024{\natexlab{a}}.

\bibitem[Huang et~al.(2024{\natexlab{b}})Huang, Zhang, Xu, He, Yu, Dong, Ma, Chanpaisit, Si, Jiang, et~al.]{huang2024vbench++}
Ziqi Huang, Fan Zhang, Xiaojie Xu, Yinan He, Jiashuo Yu, Ziyue Dong, Qianli Ma, Nattapol Chanpaisit, Chenyang Si, Yuming Jiang, et~al.
\newblock Vbench++: Comprehensive and versatile benchmark suite for video generative models.
\newblock \emph{arXiv preprint arXiv:2411.13503}, 2024{\natexlab{b}}.

\bibitem[Kazi et~al.(2014)Kazi, Chevalier, Grossman, Zhao, and Fitzmaurice]{kazi2014draco}
Rubaiat~Habib Kazi, Fanny Chevalier, Tovi Grossman, Shengdong Zhao, and George Fitzmaurice.
\newblock Draco: bringing life to illustrations with kinetic textures.
\newblock In \emph{CHI}, pages 351--360, 2014.

\bibitem[Lian et~al.(2024)Lian, Shi, Yala, Darrell, and Li]{lian2024llm}
Long Lian, Baifeng Shi, Adam Yala, Trevor Darrell, and Boyi Li.
\newblock Llm-grounded video diffusion models.
\newblock In \emph{ICLR}, 2024.

\bibitem[Lin et~al.(2023)Lin, Zala, Cho, and Bansal]{lin2023videodirectorgpt}
Han Lin, Abhay Zala, Jaemin Cho, and Mohit Bansal.
\newblock Videodirectorgpt: Consistent multi-scene video generation via llm-guided planning.
\newblock \emph{arXiv preprint arXiv:2309.15091}, 2023.

\bibitem[Liu et~al.(2024{\natexlab{a}})Liu, Zhang, Yin, and Liu]{liu2024transformer}
Jing-Yu Liu, Yan-Ming Zhang, Fei Yin, and Cheng-Lin Liu.
\newblock Transformer-based stroke relation encoding for online handwriting and sketches.
\newblock \emph{Pattern Recognition}, 148:\penalty0 110131, 2024{\natexlab{a}}.

\bibitem[Liu et~al.(2024{\natexlab{b}})Liu, Zeng, Ren, Li, Zhang, Yang, Jiang, Li, Yang, Su, et~al.]{liu2024grounding}
Shilong Liu, Zhaoyang Zeng, Tianhe Ren, Feng Li, Hao Zhang, Jie Yang, Qing Jiang, Chunyuan Li, Jianwei Yang, Hang Su, et~al.
\newblock Grounding dino: Marrying dino with grounded pre-training for open-set object detection.
\newblock In \emph{ECCV}, pages 38--55. Springer, 2024{\natexlab{b}}.

\bibitem[Lv et~al.(2024)Lv, Huang, Yan, Huang, Liu, Liu, Wen, Chen, and Chen]{lv2024gpt4motion}
Jiaxi Lv, Yi Huang, Mingfu Yan, Jiancheng Huang, Jianzhuang Liu, Yifan Liu, Yafei Wen, Xiaoxin Chen, and Shifeng Chen.
\newblock Gpt4motion: Scripting physical motions in text-to-video generation via blender-oriented gpt planning.
\newblock In \emph{CVPR}, pages 1430--1440, 2024.

\bibitem[Peebles and Xie(2023)]{peebles2023scalable}
William Peebles and Saining Xie.
\newblock Scalable diffusion models with transformers.
\newblock In \emph{ICCV}, pages 4195--4205, 2023.

\bibitem[Poole et~al.(2022)Poole, Jain, Barron, and Mildenhall]{poole2022dreamfusion}
Ben Poole, Ajay Jain, Jonathan~T Barron, and Ben Mildenhall.
\newblock Dreamfusion: Text-to-3d using 2d diffusion.
\newblock \emph{arXiv preprint arXiv:2209.14988}, 2022.

\bibitem[Qin et~al.(2024)Qin, Wu, Chen, Ren, Li, Wu, Xiao, Wang, and Wen]{qin2024diffusiongpt}
Jie Qin, Jie Wu, Weifeng Chen, Yuxi Ren, Huixia Li, Hefeng Wu, Xuefeng Xiao, Rui Wang, and Shilei Wen.
\newblock Diffusiongpt: Llm-driven text-to-image generation system.
\newblock \emph{arXiv preprint arXiv:2401.10061}, 2024.

\bibitem[Qu et~al.(2023)Qu, Wu, Fei, Nie, and Chua]{qu2023layoutllm}
Leigang Qu, Shengqiong Wu, Hao Fei, Liqiang Nie, and Tat-Seng Chua.
\newblock Layoutllm-t2i: Eliciting layout guidance from llm for text-to-image generation.
\newblock In \emph{ACM MM}, pages 643--654, 2023.

\bibitem[Rai and Sharma(2024)]{rai2024enhancing}
Gaurav Rai and Ojaswa Sharma.
\newblock Enhancing sketch animation: Text-to-video diffusion models with temporal consistency and rigidity constraints.
\newblock \emph{arXiv preprint arXiv:2411.19381}, 2024.

\bibitem[Rai et~al.(2024)Rai, Gupta, and Sharma]{rai2024sketchanim}
Gaurav Rai, Shreyas Gupta, and Ojaswa Sharma.
\newblock Sketchanim: Real-time sketch animation transfer from videos.
\newblock \emph{Computer Graphics Forum}, 43\penalty0 (8):\penalty0 e15176, 2024.

\bibitem[Smith et~al.(2023)Smith, Zheng, Li, Jain, and Hodgins]{smith2023method}
Harrison~Jesse Smith, Qingyuan Zheng, Yifei Li, Somya Jain, and Jessica~K Hodgins.
\newblock A method for animating children’s drawings of the human figure.
\newblock \emph{ACM Transactions on Graphics}, 42\penalty0 (3):\penalty0 1--15, 2023.

\bibitem[Sohn and Choy(2010)]{sohn2010sketch}
Eisung Sohn and Yoon-Chul Choy.
\newblock Sketch-n-stretch: sketching animations using cutouts.
\newblock \emph{IEEE computer graphics and applications}, 32\penalty0 (3):\penalty0 59--69, 2010.

\bibitem[Song et~al.(2021)Song, Meng, and Ermon]{song2021denoising}
Jiaming Song, Chenlin Meng, and Stefano Ermon.
\newblock Denoising diffusion implicit models.
\newblock In \emph{ICLR}, 2021.

\bibitem[Su et~al.(2018)Su, Bai, Fu, Tai, and Wang]{su2018live}
Qingkun Su, Xue Bai, Hongbo Fu, Chiew-Lan Tai, and Jue Wang.
\newblock Live sketch: Video-driven dynamic deformation of static drawings.
\newblock In \emph{CHI}, pages 1--12, 2018.

\bibitem[Su et~al.(2023)Su, Guo, Gao, Shen, and Song]{su2023motionzero}
Sitong Su, Litao Guo, Lianli Gao, Hengtao Shen, and Jingkuan Song.
\newblock Motionzero: Exploiting motion priors for zero-shot text-to-video generation.
\newblock \emph{arXiv preprint arXiv:2311.16635}, 2023.

\bibitem[Sun et~al.(2024)Sun, Huang, Liu, Wu, Xu, Li, and Liu]{sun2024t2v}
Kaiyue Sun, Kaiyi Huang, Xian Liu, Yue Wu, Zihan Xu, Zhenguo Li, and Xihui Liu.
\newblock T2v-compbench: A comprehensive benchmark for compositional text-to-video generation.
\newblock \emph{arXiv preprint arXiv:2407.14505}, 2024.

\bibitem[Suzuki et~al.(2020)Suzuki, Kazi, Wei, DiVerdi, Li, and Leithinger]{suzuki2020realitysketch}
Ryo Suzuki, Rubaiat~Habib Kazi, Li-Yi Wei, Stephen DiVerdi, Wilmot Li, and Daniel Leithinger.
\newblock Realitysketch: Embedding responsive graphics and visualizations in ar through dynamic sketching.
\newblock In \emph{UIST}, pages 166--181, 2020.

\bibitem[Tang et~al.(2024)Tang, Yang, Zhu, Zeng, and Bansal]{tang2024any}
Zineng Tang, Ziyi Yang, Chenguang Zhu, Michael Zeng, and Mohit Bansal.
\newblock Any-to-any generation via composable diffusion.
\newblock \emph{Advances in Neural Information Processing Systems}, 36, 2024.

\bibitem[Tian et~al.(2024)Tian, Yang, Yang, Gao, Deng, Chen, Wang, Yu, Tao, Wan, et~al.]{tian2024videotetris}
Ye Tian, Ling Yang, Haotian Yang, Yuan Gao, Yufan Deng, Jingmin Chen, Xintao Wang, Zhaochen Yu, Xin Tao, Pengfei Wan, et~al.
\newblock Videotetris: Towards compositional text-to-video generation.
\newblock \emph{arXiv preprint arXiv:2406.04277}, 2024.

\bibitem[Vinker et~al.(2022)Vinker, Pajouheshgar, Bo, Bachmann, Bermano, Cohen-Or, Zamir, and Shamir]{vinker2022clipasso}
Yael Vinker, Ehsan Pajouheshgar, Jessica~Y Bo, Roman~Christian Bachmann, Amit~Haim Bermano, Daniel Cohen-Or, Amir Zamir, and Ariel Shamir.
\newblock Clipasso: Semantically-aware object sketching.
\newblock \emph{ACM Transactions on Graphics}, 41\penalty0 (4):\penalty0 1--11, 2022.

\bibitem[Wang et~al.(2023)Wang, Yuan, Chen, Zhang, Wang, and Zhang]{wang2023modelscope}
Jiuniu Wang, Hangjie Yuan, Dayou Chen, Yingya Zhang, Xiang Wang, and Shiwei Zhang.
\newblock Modelscope text-to-video technical report.
\newblock \emph{arXiv preprint arXiv:2308.06571}, 2023.

\bibitem[Wang et~al.(2024)Wang, Yuan, Zhang, Chen, Wang, Zhang, Shen, Zhao, and Zhou]{wang2024videocomposer}
Xiang Wang, Hangjie Yuan, Shiwei Zhang, Dayou Chen, Jiuniu Wang, Yingya Zhang, Yujun Shen, Deli Zhao, and Jingren Zhou.
\newblock Videocomposer: Compositional video synthesis with motion controllability.
\newblock \emph{Advances in Neural Information Processing Systems}, 36, 2024.

\bibitem[Wei et~al.(2024)Wei, Shan, Zhang, Zhang, Zhang, and Ma]{wei2024real}
Yifan Wei, Wenkang Shan, Qi Zhang, Liuxin Zhang, Jian Zhang, and Siwei Ma.
\newblock Real-time interaction with animated human figures in chinese ancient paintings.
\newblock In \emph{ICMEW}, pages 1--6. IEEE, 2024.

\bibitem[Xing et~al.(2024)Xing, Xia, Zhang, Chen, Yu, Liu, Liu, Wang, Shan, and Wong]{xing2024dynamicrafter}
Jinbo Xing, Menghan Xia, Yong Zhang, Haoxin Chen, Wangbo Yu, Hanyuan Liu, Gongye Liu, Xintao Wang, Ying Shan, and Tien-Tsin Wong.
\newblock Dynamicrafter: Animating open-domain images with video diffusion priors.
\newblock In \emph{ECCV}, pages 399--417. Springer, 2024.

\bibitem[Xu et~al.(2024)Xu, Liang, Bhatt, Hu, Liang, Plataniotis, and Wang]{xu2024comp4d}
Dejia Xu, Hanwen Liang, Neel~P Bhatt, Hezhen Hu, Hanxue Liang, Konstantinos~N Plataniotis, and Zhangyang Wang.
\newblock Comp4d: Llm-guided compositional 4d scene generation.
\newblock \emph{arXiv preprint arXiv:2403.16993}, 2024.

\bibitem[Yan et~al.(2024)Yan, Zhou, Meng, Wu, Qiu, Ye, Cui, Chen, and Han]{yan2024dreamdissector}
Zizheng Yan, Jiapeng Zhou, Fanpeng Meng, Yushuang Wu, Lingteng Qiu, Zisheng Ye, Shuguang Cui, Guanying Chen, and Xiaoguang Han.
\newblock Dreamdissector: Learning disentangled text-to-3d generation from 2d diffusion priors.
\newblock In \emph{ECCV}, pages 124--141. Springer, 2024.

\bibitem[Yang et~al.(2024{\natexlab{a}})Yang, Li, Zhang, and Song]{yang2024sketchanimator}
Ruolin Yang, Da Li, Honggang Zhang, and Yi-Zhe Song.
\newblock Sketchanimator: Animate sketch via motion customization of text-to-video diffusion models.
\newblock In \emph{VCIP}, pages 1--5. IEEE, 2024{\natexlab{a}}.

\bibitem[Yang et~al.(2024{\natexlab{b}})Yang, Teng, Zheng, Ding, Huang, Xu, Yang, Hong, Zhang, Feng, et~al.]{yang2024cogvideox}
Zhuoyi Yang, Jiayan Teng, Wendi Zheng, Ming Ding, Shiyu Huang, Jiazheng Xu, Yuanming Yang, Wenyi Hong, Xiaohan Zhang, Guanyu Feng, et~al.
\newblock Cogvideox: Text-to-video diffusion models with an expert transformer.
\newblock \emph{arXiv preprint arXiv:2408.06072}, 2024{\natexlab{b}}.

\bibitem[Yin et~al.(2010)Yin, Wang, Yu, and Wang]{yin2010sketch}
Tingting Yin, Danli Wang, Kun Yu, and Hao Wang.
\newblock Sketch animation techniques and applications based on mobile devices.
\newblock In \emph{APWCS}, pages 78--81. IEEE, 2010.

\bibitem[Ying et~al.(2024)Ying, Xiong, Zhu, Qiu, Qin, and Guo]{ying2024wristsketcher}
Enting Ying, Tianyang Xiong, Gaoxiang Zhu, Ming Qiu, Yipeng Qin, and Shihui Guo.
\newblock Wristsketcher: Creating 2d dynamic sketches in ar with a sensing wristband.
\newblock \emph{International Journal of Human--Computer Interaction}, pages 1--17, 2024.

\bibitem[Yu et~al.(2023)Yu, Blackburn-Matzen, Nguyen, Wang, Habib~Kazi, and Bousseau]{yu2023videodoodles}
Emilie Yu, Kevin Blackburn-Matzen, Cuong Nguyen, Oliver Wang, Rubaiat Habib~Kazi, and Adrien Bousseau.
\newblock Videodoodles: Hand-drawn animations on videos with scene-aware canvases.
\newblock \emph{ACM Transactions on Graphics}, 42\penalty0 (4):\penalty0 1--12, 2023.

\bibitem[Zeng et~al.(2024)Zeng, Yang, Li, Liu, Zhang, Tian, Zhu, Guo, Wang, Xu, et~al.]{zeng2024trans4d}
Bohan Zeng, Ling Yang, Siyu Li, Jiaming Liu, Zixiang Zhang, Juanxi Tian, Kaixin Zhu, Yongzhen Guo, Fu-Yun Wang, Minkai Xu, et~al.
\newblock Trans4d: Realistic geometry-aware transition for compositional text-to-4d synthesis.
\newblock \emph{arXiv preprint arXiv:2410.07155}, 2024.

\bibitem[Zhang et~al.(2023)Zhang, Wang, Zhang, Zhao, Yuan, Qin, Wang, Zhao, and Zhou]{zhang2023i2vgen}
Shiwei Zhang, Jiayu Wang, Yingya Zhang, Kang Zhao, Hangjie Yuan, Zhiwu Qin, Xiang Wang, Deli Zhao, and Jingren Zhou.
\newblock I2vgen-xl: High-quality image-to-video synthesis via cascaded diffusion models.
\newblock \emph{arXiv preprint arXiv:2311.04145}, 2023.

\bibitem[Zhou et~al.(2024)Zhou, Ran, Xiong, He, Lin, Wang, Sun, and Yang]{zhou2024gala3d}
Xiaoyu Zhou, Xingjian Ran, Yajiao Xiong, Jinlin He, Zhiwei Lin, Yongtao Wang, Deqing Sun, and Ming-Hsuan Yang.
\newblock Gala3d: Towards text-to-3d complex scene generation via layout-guided generative gaussian splatting.
\newblock In \emph{ICML}, 2024.

\bibitem[Zhu et~al.(2024)Zhu, He, Tang, Guo, Chen, and Bian]{zhu2024compositional}
Hanxin Zhu, Tianyu He, Anni Tang, Junliang Guo, Zhibo Chen, and Jiang Bian.
\newblock Compositional 3d-aware video generation with llm director.
\newblock In \emph{NeurIPS}, 2024.

\end{thebibliography}
}

\newpage

\twocolumn[
\begin{center}
    {\Large \bfseries Multi-Object Sketch Animation by Scene Decomposition and Motion Planning \par}
    \vspace{2em}
    {\Large Supplementary Material \par}
\end{center}
]

\begin{figure*}[htbp]
  \centering
  \includegraphics[width=\textwidth]{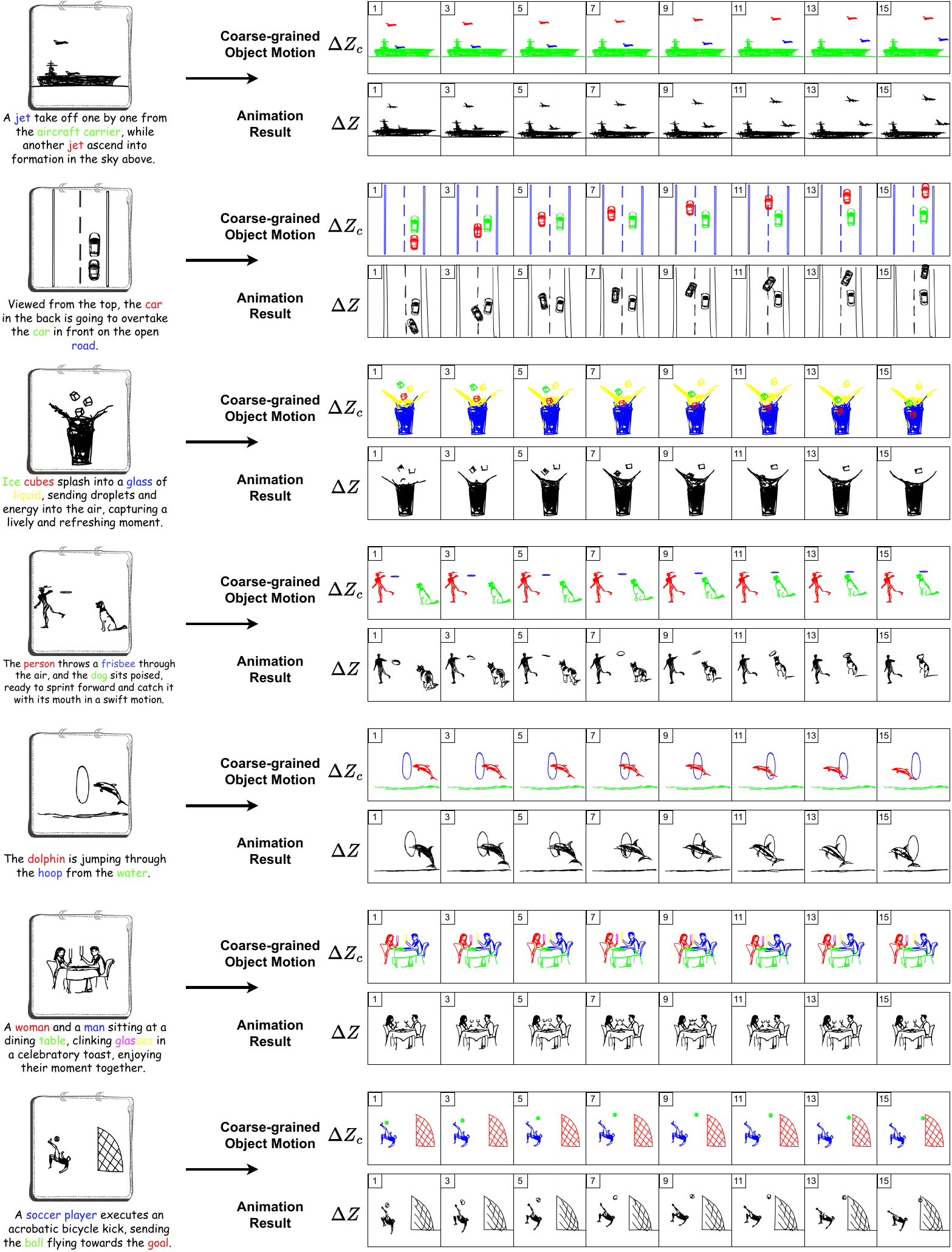}
  \caption{\textbf{More results on the created sketches}. } \label{fig:more_result}
\end{figure*}
\begin{figure*}[htbp]
  \centering
  \includegraphics[width=0.95\textwidth]{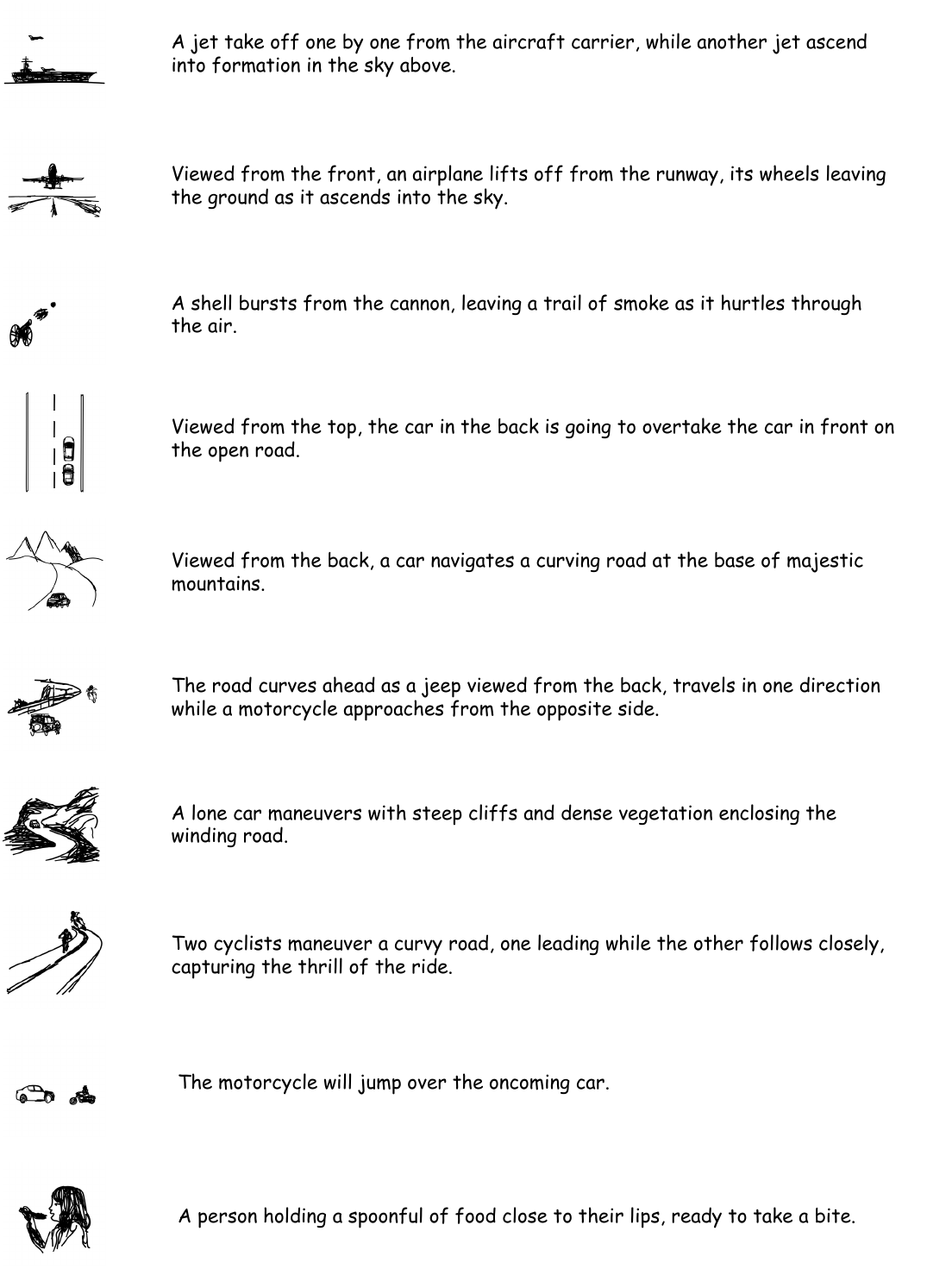}
  \caption{\textbf{The created sketches with text instructions for the ``object" class}. } \label{fig:dataset1}
\end{figure*}
\begin{figure*}[htbp]
  \centering
  \includegraphics[width=0.95\textwidth]{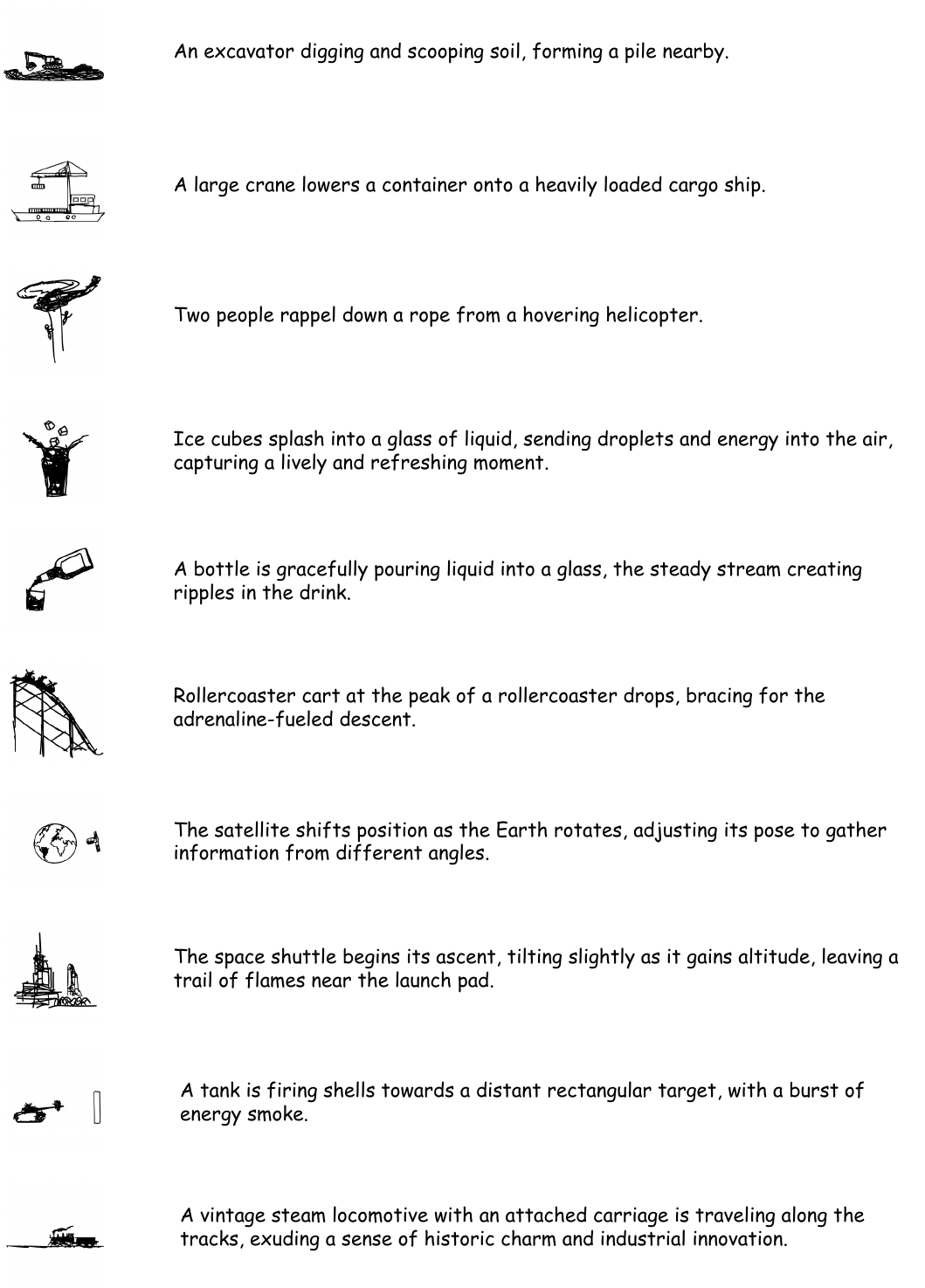}
  \caption{\textbf{The created sketches with text instructions for the ``object" class}. } \label{fig:dataset2}
\end{figure*}
\begin{figure*}[htbp]
  \centering
  \includegraphics[width=0.95\textwidth]{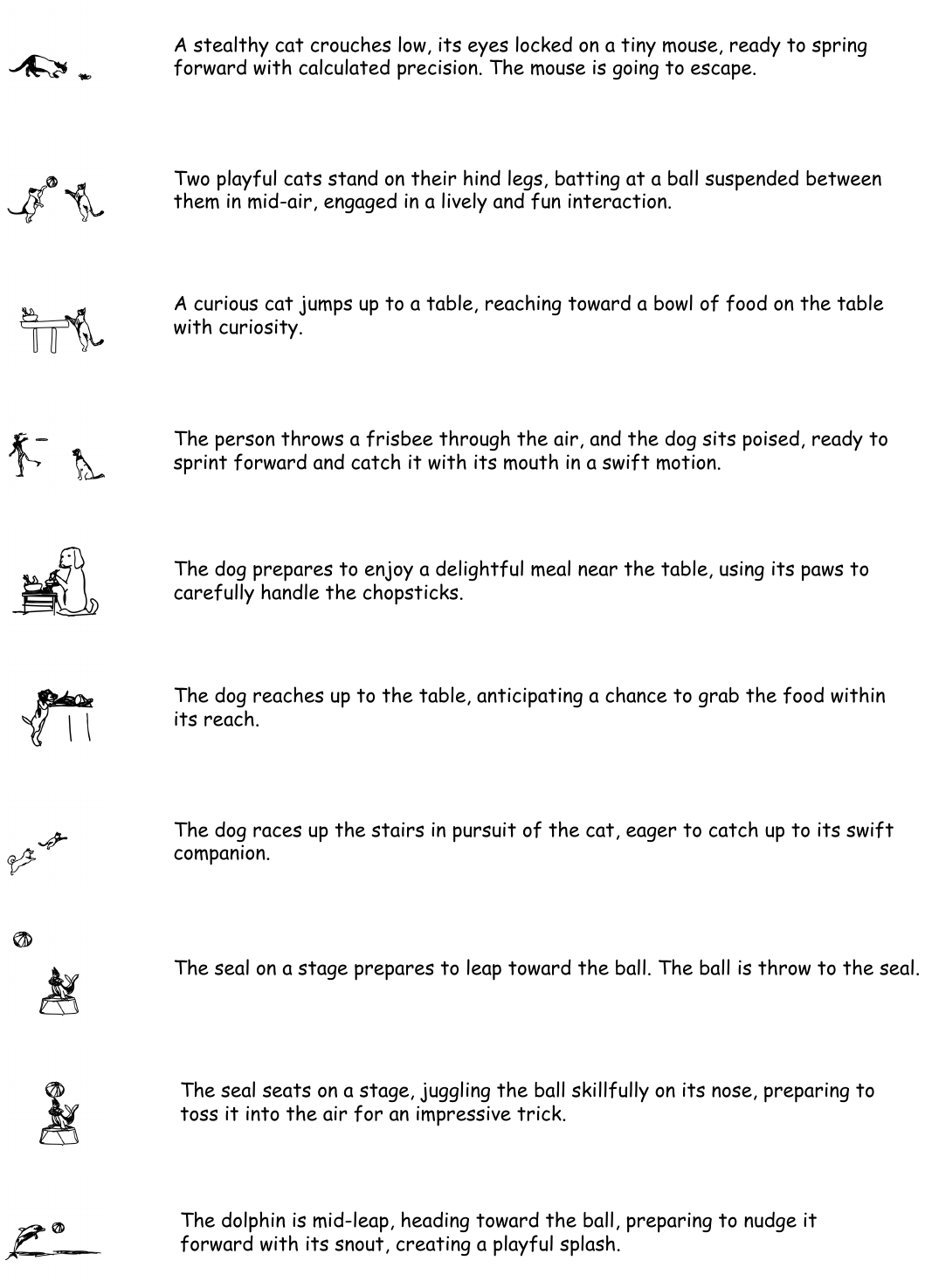}
  \caption{\textbf{The created sketches with text instructions for the ``creature" class}. } \label{fig:dataset3}
\end{figure*}
\begin{figure*}[htbp]
  \centering
  \includegraphics[width=0.95\textwidth]{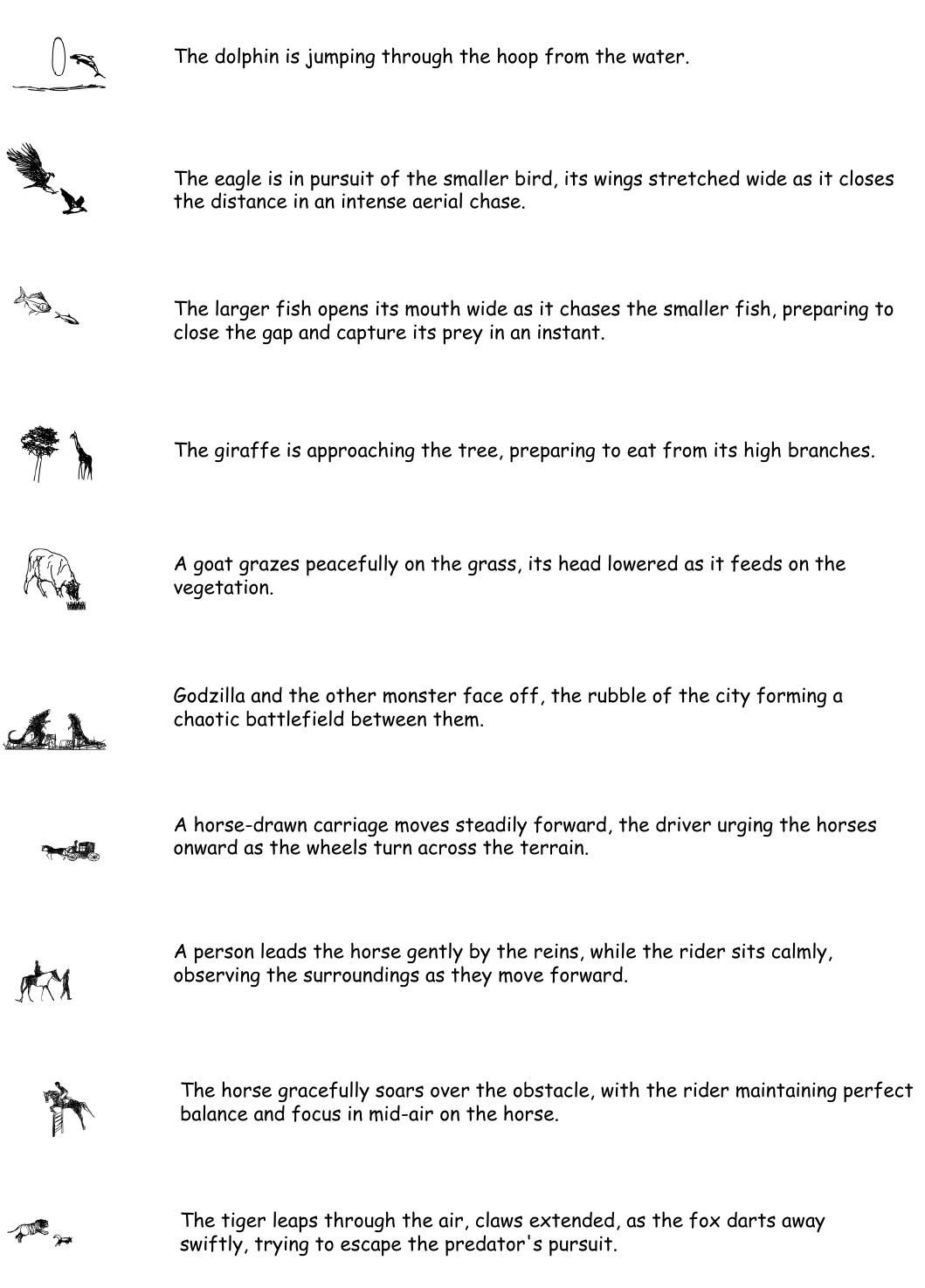}
  \caption{\textbf{The created sketches with text instructions for the ``creature" class}. } \label{fig:dataset4}
\end{figure*}
\begin{figure*}[htbp]
  \centering
  \includegraphics[width=0.95\textwidth]{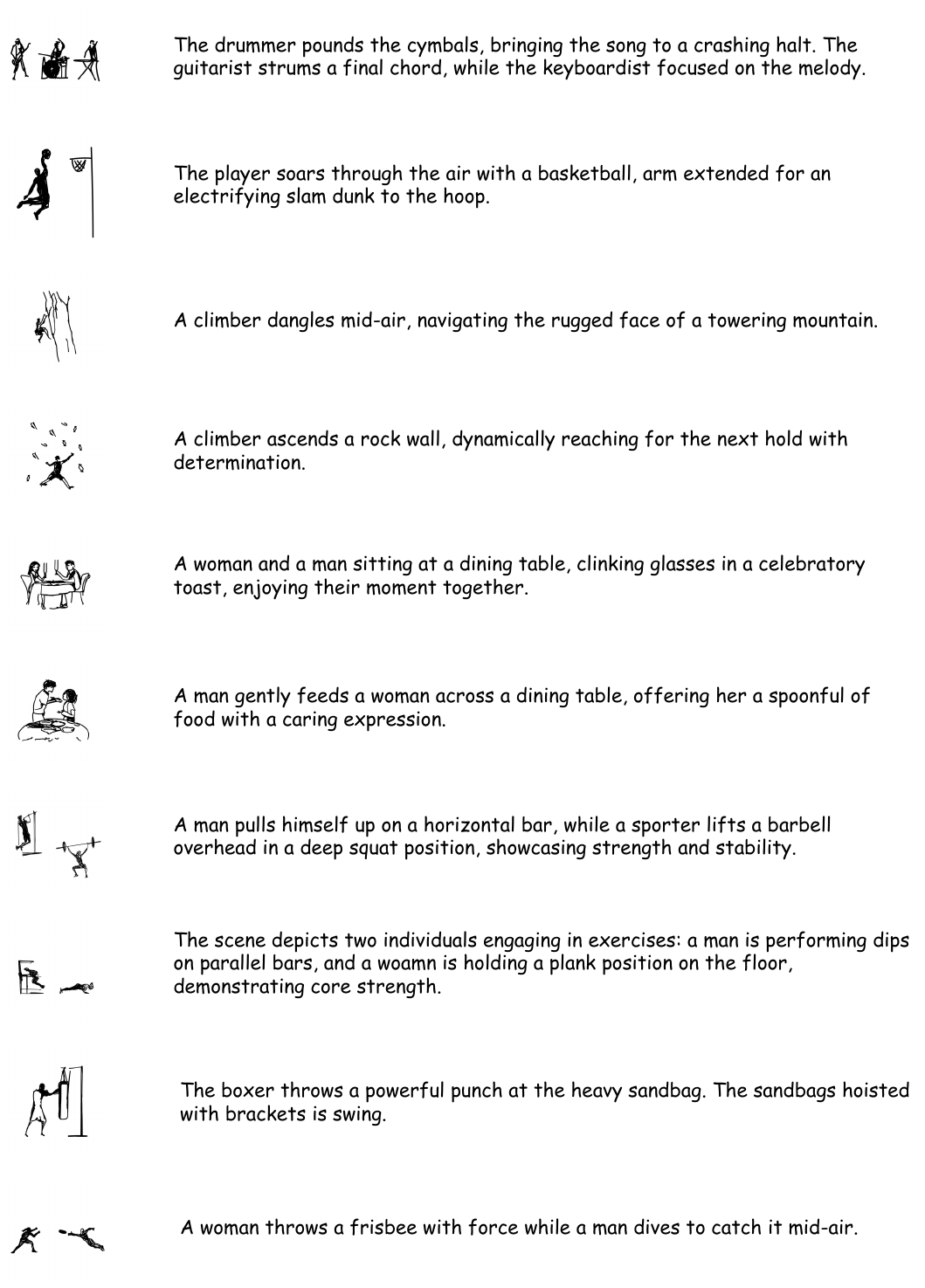}
  \caption{\textbf{The created sketches with text instructions for the ``human" class}. } \label{fig:dataset5}
\end{figure*}
\begin{figure*}[htbp]
  \centering
  \includegraphics[width=0.95\textwidth]{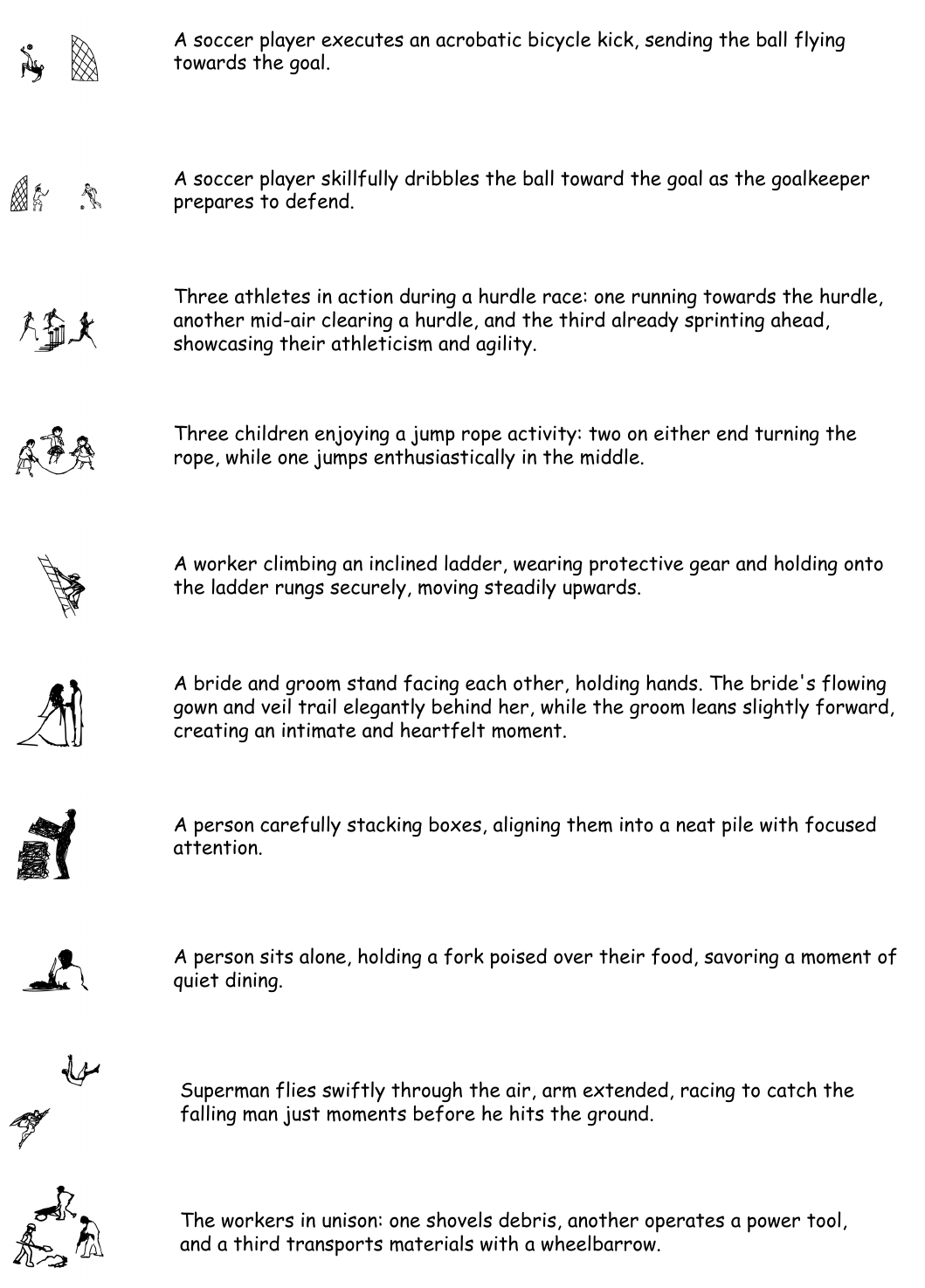}
  \caption{\textbf{The created sketches with text instructions for the ``human" class}. } \label{fig:dataset6}
\end{figure*}

\noindent In this supplementary material, we report:
\begin{itemize}
    \item More results on the created sketches (\cref{fig:more_result}).
    \item Evaluation on freehand sketches (Sec.\textcolor{iccvblue}{S1}).
    \item Evaluation on Single-object sketches (Sec.\textcolor{iccvblue}{S2}).
    \item GPT-4 instructions in MoSketch (Sec.\textcolor{iccvblue}{S3}).
    \item The created sketches with instructions (\cref{fig:dataset1,fig:dataset2,fig:dataset3,fig:dataset4,fig:dataset5,fig:dataset6}).
\end{itemize}

\section*{S1.Evaluation on Freehand Sketches} 
\begin{figure*}[htbp]
  \centering
  \includegraphics[width=\textwidth]{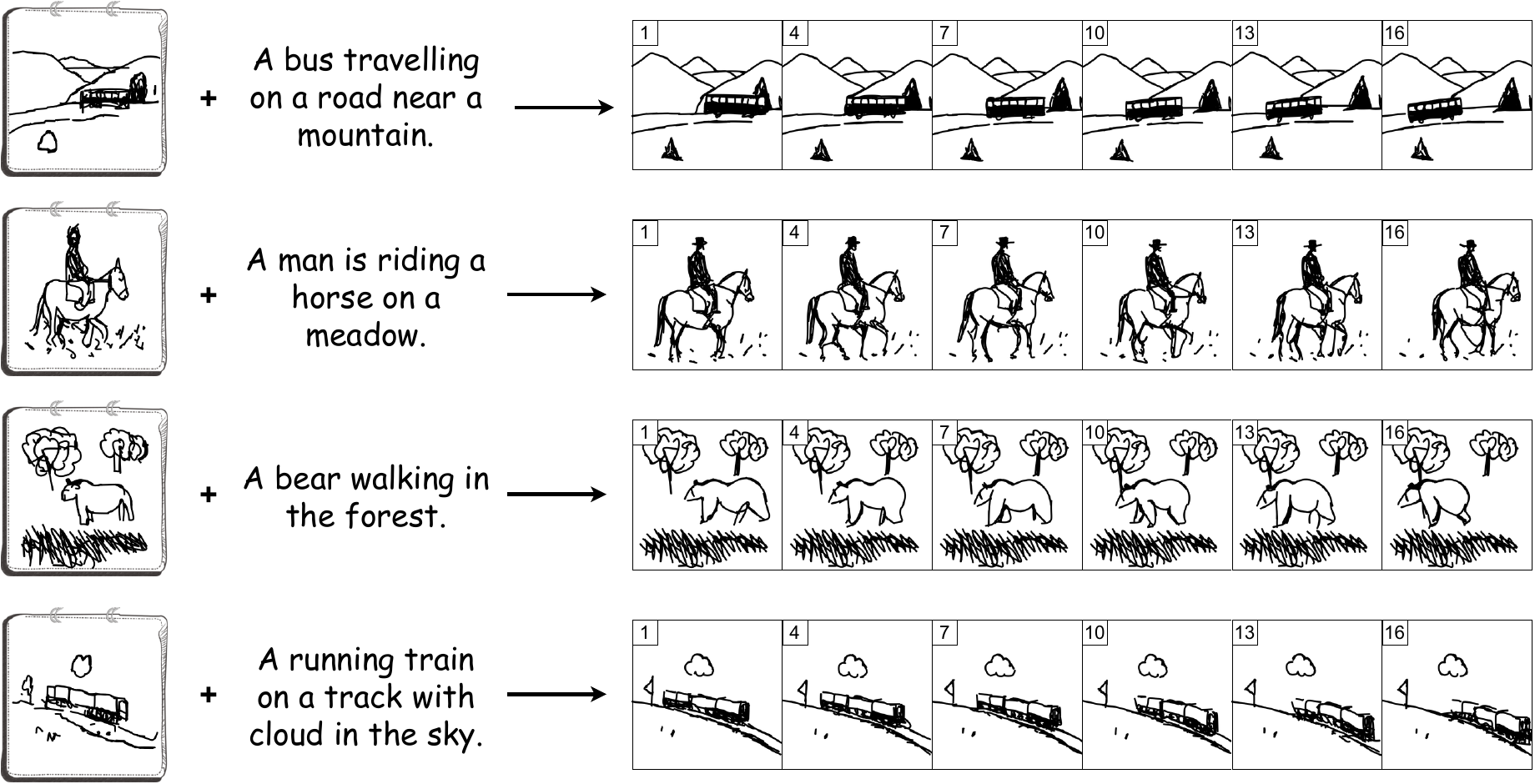}
  \caption{\textbf{Animating FS-COCO samples by MoSketch}. } \label{fig:fscoco}
\end{figure*}

To evaluate MoSketch's external validity, we selected a diverse set of 30 multi-object freehand drawings from FS-COCO~\cite{chowdhury2022fs}, a human-drawn vector scene sketch dataset. \cref{fig:fscoco} shows the animation results of MoSketch, and \cref{tab:fscoco} provides the quantitative comparison with Live-Sketch and FlipSketch. MoSketch achieves vivid multi-object sketch animation on freehand drawings, notably improving the drawing imperfections of freehand sketches.

\section*{S2.Evaluation on Single-object Sketches}
\begin{figure*}[htbp]
  \centering
  \includegraphics[width=\textwidth]{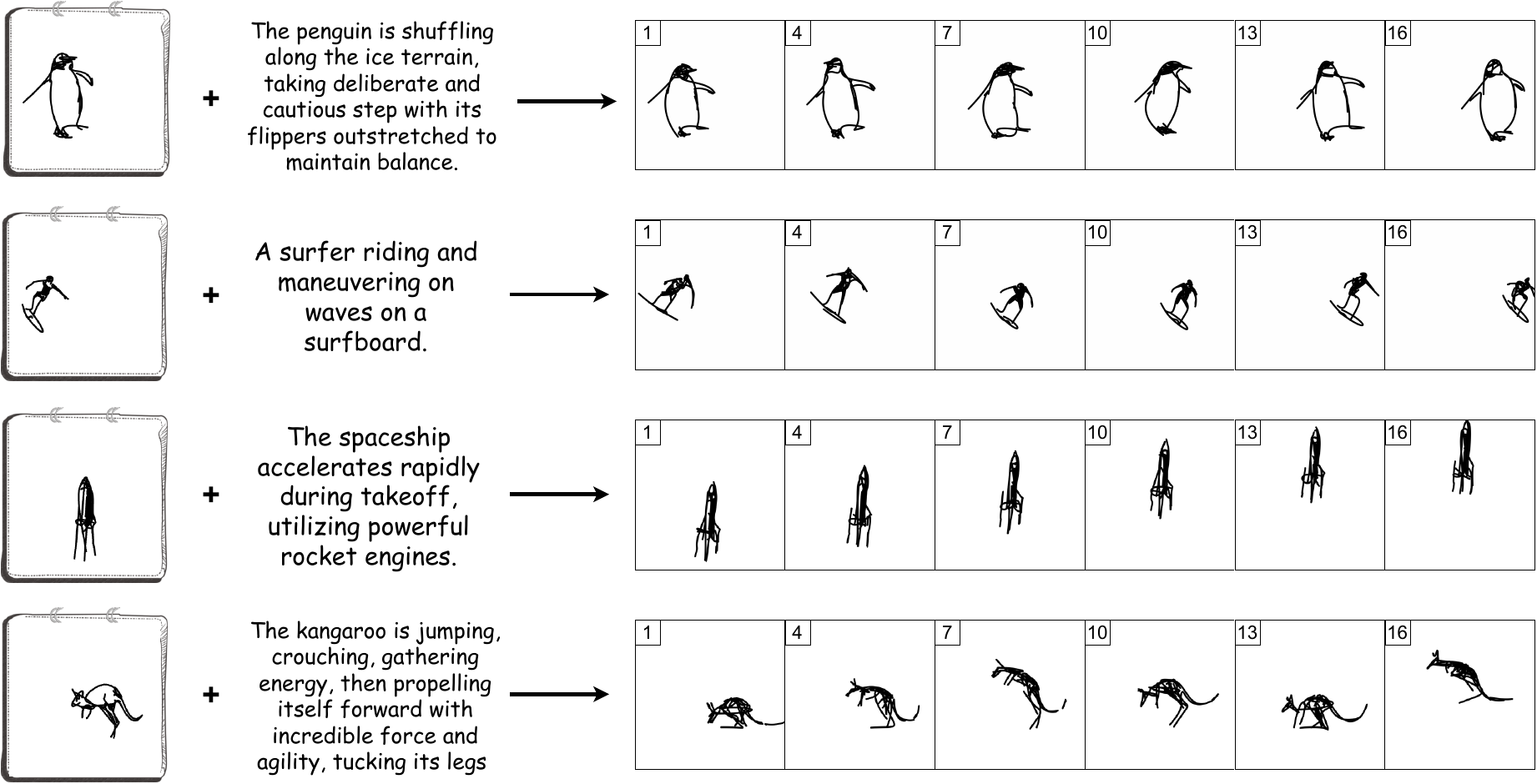}
  \caption{\textbf{Animating results on single-object sketches}. } \label{fig:single}
\end{figure*}

We evaluate MoSketch on single-object sketches from Live-Sketch, as shown in \cref{fig:single}. \cref{tab:single} shows that MoSketch is less effective in single-object scenario.

\section*{S3.GPT-4 Instructions in MoSketch}
We employ GPT-4 for scene decomposition and motion planning, and the relative instructions are listed as follows.

\subsection*{S3.1.Instructions for Scene Decomposition}
\begin{tcolorbox}[breakable]
You are an intelligent Scene Decomposition Assistant for Multi-object Sketch Animation. I will give you a sketch and a complex instruction to animate it. We want to use a divide-and-conquer method. You should decompose a complex instruction for Multi-object Sketch Animation to no more than five simple ones, and each instruction involves no more than seven objects, one or two are preferring. Objects should be used for grounding in next process, so too small and abstract objects could be ignored. Reasonable imagination is fine.

\textbf{Input:} a sketch, a complex instruction

\textbf{Output:} objects, simple instructions: [(instruction1, object\_set1),...]

\end{tcolorbox}

\subsection*{S3.2.Instructions for Motion Planning}
\begin{tcolorbox}[breakable]
You are an intelligent Motion Planning Assistant for Multi-object Sketch Animation. A sketch, an instruction to animate it and each object 's bounding box are provided. You should predict the bounding box of each object in 16 frames according to the reasonable inference. Note that the movement should follow the laws of physics such as inertia and gravity. If the sketch is in the first person, then the rule that objects far away are small and objects near are large should also be considered. Don't forget considering the interaction or relationship of objects. The image size is 256 * 256, and objects should appear in the image as far as possible.  Show me the reasoning process before planning.

\textbf{Input:} a sketch, a complex instruction, objects: [(object1,[x1, y1,w1,h1]), ...]

\textbf{Output:} the reasoning process, motion plan: $\left.\text{[(object1: [[x1,y1,w1,h1]}\right.$,...,[x16,y16,w16,h16]]),...]

\end{tcolorbox}

\begin{table}
\centering
\resizebox{\linewidth}{!}{
\begin{tabular}{lcccc}
\toprule
\multirow{2}*{\textbf{Method}} & Text-to-Video & Sketch-to-Video & Motion & Dynamic \\
                               &  Alignment & Alignment & Smoothing & Degree \\

\midrule
FlipSketch                            & 0.181 & 0.757  & 0.823  & -  \\
Live-Sketch                             & 0.173 & 0.732 & 0.827 & 0.500 \\
\midrule
MoSketch                               & \textbf{0.197} & \textbf{0.927} & \textbf{0.940} & \textbf{0.633} \\
\bottomrule
\end{tabular}}
\caption{\textbf{Evaluation on 30 freehand drawings from FS-COCO}.}
\label{tab:fscoco}
\end{table} 

\begin{table}
\centering
\resizebox{\linewidth}{!}{
\begin{tabular}{lcccc}
\toprule
\multirow{2}*{\textbf{Method}} & Text-to-Video & Sketch-to-Video & Motion & Dynamic \\
                               &  Alignment & Alignment & Smoothing & Degree \\

\midrule
FlipSketch                            & 0.211 & \textbf{0.936}  & 0.858  & \textbf{1.000}  \\
Live-Sketch                             & \textbf{0.217} & 0.884 & 0.827 & 0.392 \\
\midrule
MoSketch                               & 0.209 & 0.914 & \textbf{0.968} & 0.571  \\
\bottomrule
\end{tabular}}
\caption{\textbf{Performance comparison in a single-object scenario}.}
\label{tab:single}
\end{table}

\end{document}